\documentclass[runningheads]{llncs}
\usepackage{graphicx}
\usepackage{microtype}
\usepackage{graphicx}
\usepackage{booktabs} % for professional tables
\usepackage{makecell}
\usepackage{multirow}
\usepackage{algorithmic}
\usepackage{algorithm}
\usepackage{amsmath}
\usepackage{amssymb}
\usepackage{mathtools}
\usepackage{xcolor}
\usepackage{dsfont}
\usepackage{wrapfig}
\usepackage[export]{adjustbox}
\usepackage{url}

\usepackage{subcaption}

\usepackage{amsmath,amsfonts,bm}

\def\eqref#1{equation~\ref{#1}}
\def\1{\bm{1}}

\def\rx{{\textnormal{x}}}

\def\rz{{\textnormal{z}}}

\def\rvx{{\mathbf{x}}}

\def\rvz{{\mathbf{z}}}

\DeclareMathAlphabet{\mathsfit}{\encodingdefault}{\sfdefault}{m}{sl}
\SetMathAlphabet{\mathsfit}{bold}{\encodingdefault}{\sfdefault}{bx}{n}

\newcommand{\E}{\mathbb{E}}

\newcommand{\KL}[2]{D_{\mathrm{KL}}\left[#1\|#2\right]}

\begin{document}

\title{Learning Data Representations\\ with Joint Diffusion Models}
% \author{Anonymous Author}
\author{Kamil Deja\inst{1,2}\orcidID{0000-0003-1156-5544} \and
Tomasz Trzcinski\inst{1,2}\orcidID{0000-0002-1486-8906} \and
Jakub M. Tomczak\inst{3}\orcidID{0000-0001-8634-694X}}
\authorrunning{K. Deja et al.}
% \institute{Anonymous Institute}
\institute{Warsaw University of Technology, Warsaw, Poland\\
\email{kamil.deja@pw.edu.pl}, \email{tomasz.trzcinski@pw.edu.pl} \and
IDEAS NCBR, Warsaw, Poland \and
Eindhoven University of Technology, Eindhoven, The Netherlands\\
\email{j.m.tomczak@tue.nl}}

\maketitle              % typeset the header of the contribution
\begin{abstract}
% We introduce a joint diffusion model that simultaneously learns meaningful internal representations fit for both generative and predictive tasks. 
% Joint machine learning models that allow synthesizing and classifying data often offer uneven performance between those tasks or are unstable to train. In this work, we depart from a set of empirical observations that indicate the usefulness of internal representations built by contemporary deep diffusion-based generative models in both generative and predictive settings \kamil{It sounds a bit as if diffusion models were build in generative and predictive settings}. We then introduce an extension of the vanilla diffusion model with a classifier that allows for stable joint training with shared parametrization between those objectives. The resulting joint diffusion model outperforms recent state-of-the-art hybrid methods in terms of both classification and generation quality on all evaluated benchmarks. On top of our joint training approach, we present how we can directly benefit from shared generative and discriminative representations by introducing a method for visual counterfactual explanations.
Joint machine learning models that allow synthesizing and classifying data often offer uneven performance between those tasks or are unstable to train. In this work, we depart from a set of empirical observations that indicate the usefulness of internal representations built by contemporary deep diffusion-based generative models not only for generating but also predicting. We then propose to extend the vanilla diffusion model with a classifier that allows for stable joint end-to-end training with shared parameterization between those objectives. The resulting joint diffusion model outperforms recent state-of-the-art hybrid methods in terms of both classification and generation quality on all evaluated benchmarks. On top of our joint training approach, we present how we can directly benefit from shared generative and discriminative representations by introducing a method for visual counterfactual explanations.
% On top of our joint training approach, we introduce a method for visual counterfacactual explanations. % offers superior performance in across various tasks, including generative modeling, classification, and visual counterfactual explanations.
\keywords{Deep generative models, diffusion models, joint models}
\end{abstract}
\section{Introduction}
\label{sec:Introduction}

Training a single machine learning model that can jointly %\textit{(i)}~
synthesize new data as well as %\textit{(ii)}~
to make predictions about input samples remains a long-standing goal of machine learning~\cite{jebara2012machine,lasserre2006principled}. 
Shared representations created with a combination of those two objectives promise benefits on many downstream problems such as calibration of model uncertainty~\cite{chapelle2009semi}, semi-supervised learning~\cite{kingma2014semi}, unsupervised domain adaptation~\cite{ilse2020diva} or continual learning~\cite{masarczyk2021onrob}.

Therefore, since the introduction of deep generative models such as Variational Autoencoders (VAEs)~\cite{kingma2014autoencoding}, a growing body of work takes advantage of shared deep neural network-based parameterization and latent variables to build joint models.
For instance, \cite{ilse2020diva,tulyakov2017hybrid,knop2020cramer,yang2022chroma} stack a classifier on top of latent variables sampled from a shared encoder. Similarly, \cite{nalisnick2019hybrid,perugachi2021invertible} use normalizing flows to obtain an invertible representation that is further fed to a classifier. However, these approaches require modifying the log-likelihood function by scaling either the conditional log-likelihood or the marginal log-likelihood. This idea, known as hybrid modeling~\cite{lasserre2006principled}, leads to the situation where models concentrate either on synthesizing data or predicting but not on both of those tasks simultaneously. 

% \begin{figure}
%     \centering
%     \includegraphics[width=\linewidth]{Figures/teaser.pdf}
%     \caption{The overview of our method. We propose to jointly train the diffusion model and the classifier using a single parametrisation with a shared UNet architecture.}
%     \label{fig:teaser}
% \end{figure}

We address existing joint models' limitations and leverage the recently introduced diffusion-based deep generative models (DDGM)~\cite{dhariwal2021diffusion,kingma2021variational,sohl2015deep}. This new family of methods has become  popular because of the unprecedented quality of the samples they generate. However, relatively little attention was paid to their inner workings, especially to the internal representations built by the DDGMs. In this work, we fill this gap and empirically analyze those representations, validating their usefulness for predictive tasks and beyond. Then, we introduce a joint diffusion model, where a classifier shares the parametrization with the UNet encoder by operating on the extracted latent features. 
This results in meaningful data representations shared across discriminative and generative objectives. 

We validate our approach in several use cases where we show how one part of our model can benefit from the other. First, we investigate how DDGMs benefit from the additional classifier to conditionally generate new samples or alter original images. Next, we show the performance improvement our method brings in the classification task. Finally, we present how we can directly benefit from joint representations used by both the classifier and generator by creating visual counterfactual explanations, namely, how to explain decisions of a model by identifying which regions of an input image need to change in order for the system to produce a specified output.
% Finally, we extend the evaluation of our joint diffusion model to semi-supervised learning, domain adaptation, and counterfactual explanations. 
% For all of those tasks, our method does not require any problem-specific adjustments, which confirms the flexibility of our approach. 

We can summarize the contributions of our work as follows:
\begin{itemize}
\item We provide empirical observations with insights into representations built internally by diffusion models, on top of which we introduce a joint classifier and diffusion model with shared parametrization.
\item We introduce a conditional sampling algorithm where we optimize internal diffusion representations with a classifier.
\item We present state-of-the-art results in terms of joint modeling where our solution outperforms other joint and hybrid methods in terms of both quality of generations and the classification performance. %prove that our solution work with several use cases including data synthesis, classification, and visual counterfactual explanation.
\end{itemize}

%===SubSECTION===
\section{Background}
% In this section, we provide a brief introduction into diffusion models. 
\textbf{Joint models} 
Let us consider two random variables: $\rvx \in \mathcal{X}$ and $y \in \mathcal{Y}$. For instance, in the classification problem we can have $\mathcal{X} = \mathbb{R}^{D}$ and $\mathcal{Y} = \{0, 1, \ldots, K-1\}$. The joint distribution over these random variables could be factorized in one of the following two manners, namely, $p(\rvx, y) = p(\rvx | y) p(y) = p(y|\rvx) p(\rvx)$. 
% \begin{align}
% p(\rvx, y) &= p(\rvx | y) p(y)\ \label{eq:p_x_y_p_y}\\
% &= p(y|\rvx)\ p(\rvx) . \label{eq:p_y_x_p_x}
% \end{align}
% In Eq. (\ref{eq:p_y_x_p_x}), we get the conditional distribution
Following the second factorization gives us the conditional distribution $p(y|\rvx)$ (e.g., a classifier) and the marginal distribution $p(\rvx)$. For prediction, it is enough to learn the conditional distribution, which is typically parameterized with neural networks.
However, training the joint model with shared parametrization has many advantages since one part of the model can positively influence the other.

\textbf{Diffusion-based Deep Generative Models} 
In this work, we follow the formulation of Diffusion-based deep generative models as presented in~\cite{ho2020denoising,sohl2015deep}. Given a data distribution $\rvx_0 \sim q(\rvx_0)$, we define a \textit{forward} noising process $q$ that produces a sequence of latent variables $\rx_1$ through $\rx_T$ by adding Gaussian noise at each time step $t$, with a variance of $\beta_t \in (0,1)$, defined by a schedule $\beta_1,...,\beta_T$, namely, $q(\rvx_1, \ldots , \rvx_{T} | \rvx_{0}) = \prod_{t=1}^{T} q(\rvx_t | \rvx_{t-1})$, where $q(\rvx_t|\rvx_{t-1}) = \mathcal{N}(\rvx_t; \sqrt{1 - \beta_t} \rvx_{t-1}, \beta_t \mathbf{I})$. 

Following~\cite{huang2021variational,kingma2021variational,tomczak2022deep,tzen2019neural}, we consider DDGMs as infinitely deep hierarchical VAEs with a specific family of variational posteriors; namely, Gaussian diffusion processes \cite{sohl2015deep}. Therefore, for data point $\rvx_0$, and latent variables $\rvx_{1}, \ldots, \rvx_{T}$, we want to optimize the marginal likelihood $p_{\theta}(\rvx_0) = \int p_{\theta}(\rvx_0, \ldots, \rvx_{T}) \mathrm{d} \rvx_1, \ldots, \rvx_{T}$, where $p_{\theta}(\rvx_0, \ldots, \rvx_{T}) = p(\rvx_T)\ \prod_{t=0}^{T} p_{\theta}(\rvx_{t-1} | \rvx_{t})$ is the \textit{backward} diffusion process with $p_{\theta}(\rvx_{t-1} | \rvx_{t}) = \mathcal{N}(\rvx_{t-1}; \mu_{\theta}(\rvx_{t}, t), \Sigma_{\theta}(\rvx_{t}, t))$.

We can define the variational lower bound as follows:
\begin{align} \label{eq:diff_elbo}
    \ln p_{\theta}(\rvx_0) \geq L_{vlb}(\theta) :=& \underbrace{\E_{q(\rvx_1|\rvx_0)}[\ln p_{\theta}(\rvx_0|\rvx_1)]}_{-L_0} - \underbrace{\KL{q(\rvx_T |\rvx_0)}{p(\rvx_T)}}_{L_T} + \notag \\
    &- \sum_{t=2}^T \underbrace{\E_{q(\rvx_t|\rvx_0)} \KL{q(\rvx_{t-1}|\rvx_t, \rvx_0)}{p_{\theta}(\rvx_{t-1}|\rvx_t)}}_{L_{t-1}}.
\end{align}
that we further optimize with respect to the parameters of the backward diffusion.

\textbf{Training objective} 
The authors of \cite{ho2020denoising} notice that instead of estimating the probability of previous latent variable $p(\rvx_{t-1}|\rvx_t)$, we can predict the added noise $\epsilon$. Therefore, a single part of the variational lower bound is equal to:
\begin{equation}\label{eq:l_t}
    L_{t}(\theta) = \mathbb{E}_{\rvx_{0}, \boldsymbol{\epsilon}}\left[\frac{\beta_{t}^{2}}{2 \sigma_{t}^{2} \alpha_{t}\left(1-\overline{\alpha}_{t}\right)} \left\|\boldsymbol{\epsilon} -\boldsymbol{\epsilon}_{\theta}\left(\sqrt{\overline{\alpha}_{t}} \rvx_{0}+\sqrt{1-\overline{\alpha}_{t}} \boldsymbol{\epsilon}, t\right)\right\|^{2}\right] ,
\end{equation}
where $\boldsymbol{\epsilon} \sim \mathcal{N}(\mathbf{0}, \mathbf{I})$ and $\boldsymbol{\epsilon}_\theta(\cdot, \cdot)$ is a neural network predicting the noise $\boldsymbol{\epsilon}$ from $\rvx_t$.

In \cite{ho2020denoising}, it is also suggested to train the model with a simplified objective that is a modified version of equation (\ref{eq:l_t}) without scaling, namely:
\begin{equation}\label{eq:l_t_simple}
% \resizebox{0.91\linewidth}{!}{$
    L_{t,\text {simple}}(\theta) = \mathbb{E}_{\mathbf{x}_{0}, \boldsymbol{\epsilon}}\left[\left\|\boldsymbol{\epsilon}-\boldsymbol{\epsilon}_{\theta}\left(\sqrt{\overline{\alpha}_{t}} \mathbf{x}_{0}+\sqrt{1-\overline{\alpha}_{t}} \boldsymbol{\epsilon}, t\right)\right\|^{2}\right] .
    % $}
\end{equation}
In practice, a single shared neural network is used for modeling $\boldsymbol{\epsilon}_{\theta}$. For that end, most of the works~\cite{ho2020denoising,kingma2021variational,nichol2021improved} use UNet architecture~\cite{ronneberger2015u} that can be seen as a specific type of an autoencoder. This is particularly relevant for this work since we benefit from the \emph{Encoder} -- \emph{Decoder} structure of the denoising DDGM model.

%===SubSECTION===
\section{Related Work}
\label{sec:related}

\textbf{Diffusion models} There are several extensions to the baseline DDGM setup that aim to improve the quality of sampled generations~\cite{ho2020denoising,huang2021variational,kingma2021variational,song2019generative,song2020score}. %This includes noising and sampling schedulers and modified training objectives. 
% \paragraph{Guided diffusion}
Several works propose to improve the quality of samples from DDGMs by conditioning the generations with class identities~\cite{ho2022classifierfree,huang2022improving,NEURIPS2021_cfe8504b}. Among those works, \cite{dhariwal2021diffusion} introduces a classifier-guided generation, where a gradient from an externally and independently trained classifier is added in the process of backward diffusion to guide the generation towards a target class. On top of this approach, \cite{augustin2022diffusion} present a tool for investigating the decision of a classifier by generating visual counterfactual explanations with a diffusion model. In this work, we simplify both of those methods benefiting from training a joint model with representations shared between a diffusion model and a classifier.

\textbf{Diffusion models and UNet representations} In \cite{abstreiter2021diffusionbased} additional encoded information to the score estimator is introduced, which allows using the score matching loss function for learning data representations. The authors of \cite{baranchuk2021labelefficient} use activations from the pre-trained diffusion UNet model for the image segmentation task. Here, we first analyze how pre-trained models could be useful for classification, and further propose a joint model that is trained end-to-end with generative and discriminative losses. Other works consider data representations from the UNet model within other generative models, e.g., a conditional UNet-based variational autoencoder \cite{esser2018variational}. Additionally in \cite{falck2022a} authors show the connection between the UNet architecture and wavelet transformation, applying it to the hierarchical VAEs. In this work, we show that indeed diffusion models learn useful representations, and further take advantage of that fact in a shared parameterization between a diffusion model and a classifier in a joint model. 

\textbf{Joint training} Apart from latent variable joint models, in \cite{grathwohl2019your} authors show that it is possible to use a shared parameterization (a neural network-based classifier) to formulate an energy-based model. This Joint Energy-based Model (JEM) could be seen as a classifier if a softmax function is applied to logits or a generator if a Markov-chain Monte Carlo method is used to sample from the model. Although it obtains strong empirical results, gradient estimators used to train JEM are unstable and prone to diverging when optimization parameters are not perfectly tuned, which limits the robustness and applicability of this method. Alternatively, Introspective Neural Networks could be used for generative modeling and classification by applying a single parameterization~\cite{jin2017introspective,lazarow2017introspective,lee2018wasserstein}. The idea behind this class of models relies on utilizing a training procedure that combines adversarial learning and contrastive learning. Similarly to JEMs, sampling is carried out by running an MCMC method. In \cite{grathwohl2021no}, the authors improve the performance of JEM by introducing a variational-based approximator (VERA) instead of MCMC. Similarly, in \cite{yang2021jem++} authors introduce JEM++, an improvement over the JEM's generative performance by applying a proximal SGLD-based generation, and classification accuracy with informative initialization. From a conceptually different perspective, the authors of \cite{yang2022your} propose an implementation of a joint model based on the Vision Transformer~\cite{dosovitskiy2020image} architecture, that yields state-of-the-art result in terms of image classification. %depart from the currently best performing Vision Transformer~\cite{dosovitskiy2020image} %model and turn it into a joint one with a generative part implemented as a diffusion process. 
Here, we propose to combine standard diffusion models with classifiers by sharing their parameterization. Thus, our training is entirely based on the log-likelihood function and end-to-end, while sampling is carried out by backward diffusion instead of any MCMC algorithm. %\kamil{I'm not sure if now it's ok - it steal sounds a bit like if we didn't do much}

%===SECTION===
\section{Diffusion models learn data representations}
\label{sec:diffusion_representations}
Learning useful data representations is important for having a good generator or classifier. Ideally, we would like to train a joint model that allows us to obtain proper representations for both $p(y|\rvx)$ and $p(\rvx)$ simultaneously. In this work, we investigate parameterizations of DDGMs and, in particular, the use of an autoencoder as a denoising decoder $p_\theta(\rvx_{t-1}|\rvx_t)$. Within this architecture, the denoising function can be decomposed into two parts: encoding of the image at the current timestep into a set of features $\mathcal{Z}_t = e(\rvx_{t})$ and then decoding it to obtain $\rvx_{t-1} = d(\mathcal{Z}_t)$. 

\begin{wrapfigure}{r}{0.6\textwidth}
    \vskip -12mm
    \begin{adjustbox}{center}
    \includegraphics[width=\linewidth]{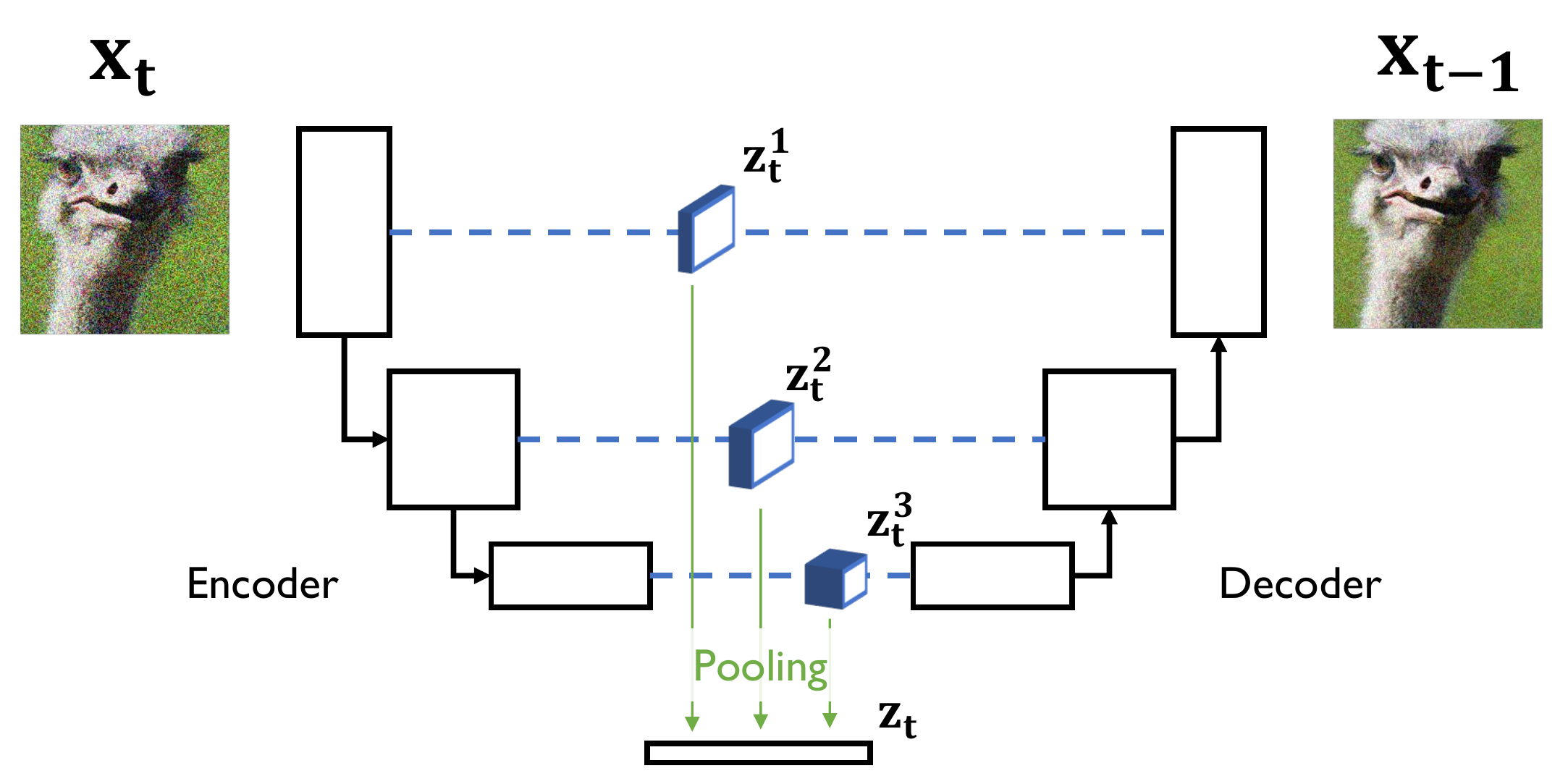}
    \end{adjustbox}
    \caption{Data representation $\rz_t$ in a UNet-based diffusion model.}
    \label{fig:unet_representation}
    \vskip -7mm
\end{wrapfigure}

For the UNet architecture, a set of features obtained from an input is a structure composed of several tensors with image features encoded to different levels, $\mathcal{Z}_t = \{\rvz_t^1, \rvz_t^2 \dots \rvz_t^n \} $. For all further experiments, we propose to pool features encoded by the same filter and concatenate the averaged representations into a single vector $\rvz_t$, as presented in Fig.~\ref{fig:unet_representation} for $n=3$. In particular, we can use average pooling to select average convolutional filter activations to the whole input. Details of this procedure are described in the Appendix~\ref{app:pooling_details}.

%===SubSECTION===
\subsection{Diffusion model representations are useful for prediction}

\begin{figure}[h]
    \vskip -5mm
     \centering
     \begin{subfigure}[b]{0.49\textwidth}
         \centering
         \includegraphics[width=\textwidth]{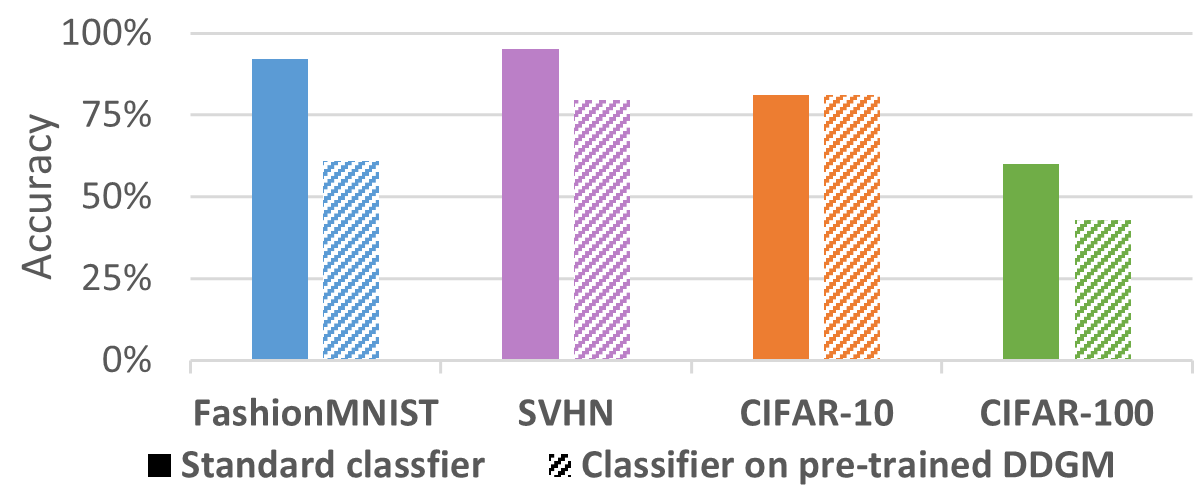}
         \caption{}
         \label{fig:accuracy_barplot}
     \end{subfigure}
     \hfill
     \begin{subfigure}[b]{0.49\textwidth}
         \centering
         \includegraphics[width=\textwidth]{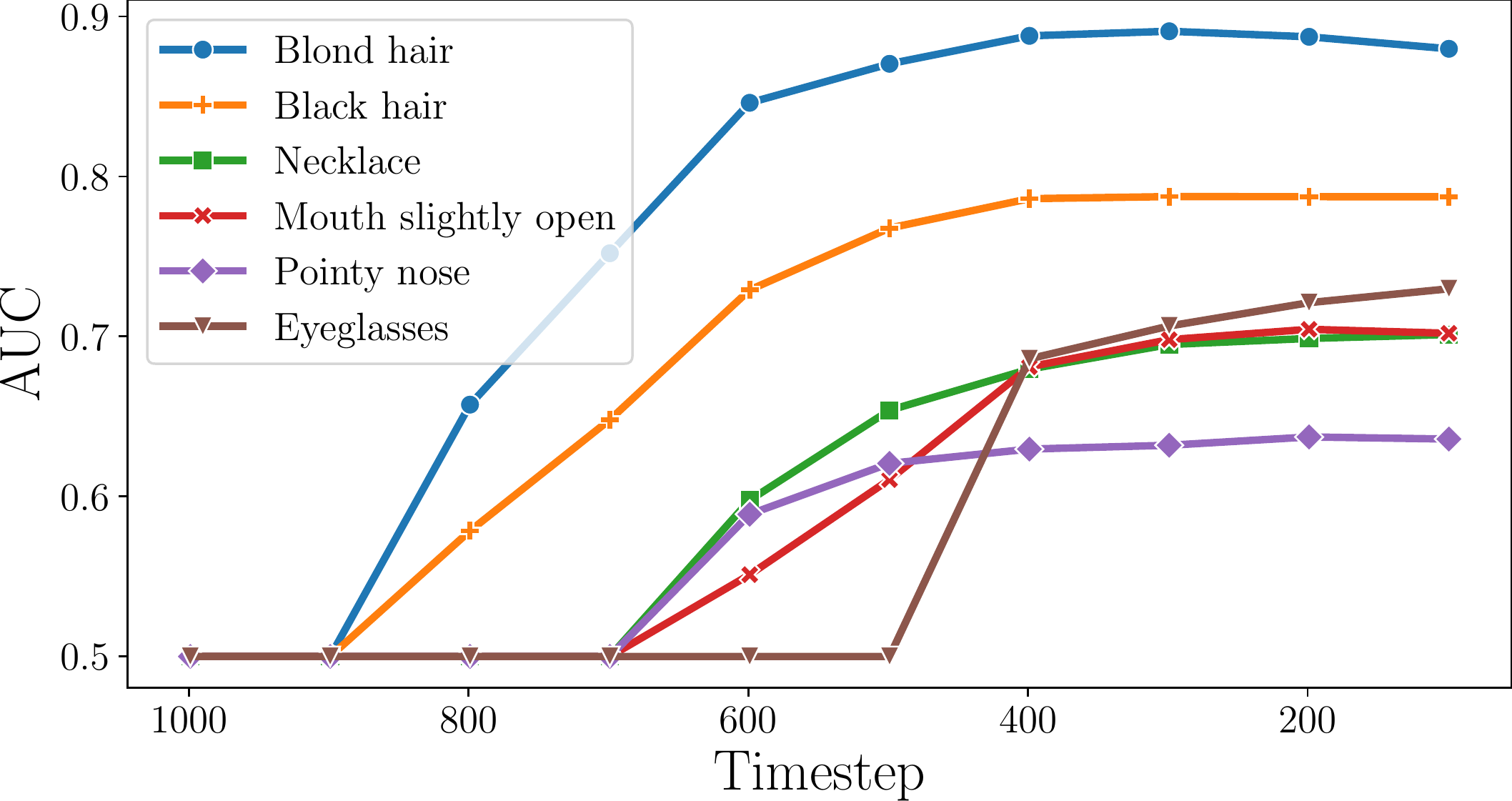}
         \caption{ }
         \label{fig:celeba_attributes}
     \end{subfigure}
        \caption{(a) The test-set accuracy of a stand-alone classifier compared to a classifier trained on top of data representations from a pre-trained diffusion model extracted from original images $\rvx_0$. (b) The area under the ROC curve (AUC) for logistic regression models fit on data representations extracted with a pre-trained diffusion model at ten different diffusion timesteps. High-grained features are already distinguishable at late diffusion steps (closer to random noise), while low-grained features are only represented at the earlier stage of the forward diffusion.}
        \label{fig:representations}
        \vskip -7mm
\end{figure}

First, we verify whether averaged representations $\rvz_0$ extracted from an original image $\rvx_0$ by the UNet contain information that is in some sense predictive. We measure it with the classification accuracy of an MLP-based classifier fed with $\rvz_0$. As presented in Fig~\ref{fig:accuracy_barplot}, representations encoded in $\rvz_0$ are indeed very informative and, in some cases (e.g., CIFAR-10), could lead to performance comparable to a stand-alone classifier with the same architecture as the combination of the UNet encoder and MLP but trained with the cross-entropy loss function. A similar observation was made in \cite{baranchuk2021labelefficient}, where the pre-trained diffusion model was used for semantic image segmentation. %\kamil{Tu nie wiem czy sie nie podkladamy :D} %This is especially true for more complex datasets such as CIFAR-10 or CIFAR-100, where feature generalization to unseen test data samples is more challenging.

% % \begin{wrapfigure}{r}{0.6\textwidth}
% \begin{figure}
% \centering
% % \vskip  -10pt
%     % \begin{adjustbox}{center}
%     \includegraphics[width=0.75\linewidth]{Figures/classifier_accuracy_barplot.png}
%     % \end{adjustbox}
%     % \vskip -5pt
%     \caption{The test-set accuracy of a stand-alone classifier compared to a classifier trained on top of data representations from a pre-trained diffusion model extracted from original images $\rvx_0$.}
%     \label{fig:accuracy_barplot} 
%     % \vskip  -5pt
%     \end{figure}
% % \end{wrapfigure}

%===SubSECTION===
\subsection{Diffusion models learn features of increasing granularity}\label{sect:diffusion_learn_features}
The next question is how the data representations $\rvz_t$ differ with diffusion timesteps $t$. To investigate this issue, we train an unsupervised DDGM on the CelebA dataset, which we then use to extract the features $\rvz_t$ at different timesteps. On top of those representations, we fit a binary logistic regression classifier for each of the 40 attributes in the dataset. In Fig.~\ref{fig:celeba_attributes}, we show the performance of those regression models for $6$ different attributes when calculated on top of representations from ten different diffusion timesteps. We observe that the model learns different data features depending on the amount of noise added to the original data sample. As presented in Fig.~\ref{fig:celeba_attributes}, high-grained data features such as hair color start to emerge at late diffusion steps (closer to the noise), while low-grained features (e.g., necklace or glasses) are not present until the early steps. This observation is in line with the works on denoising autoencoders where authors observe similar behavior for denoising with different amounts of added noise \cite{chandra2014adaptive,geras2014scheduled,zhang2018convolutional}.

% % \begin{wrapfigure}{r}{0.6\textwidth}
% \begin{figure}[!htbp]
%     \centering
%     % \begin{adjustbox}{center}
%     \includegraphics[width=0.8\linewidth]{Figures/CelebA_attributes_timesteps.pdf}
%     % \end{adjustbox}
%     % \vskip -2mm
%     \caption{The area under the ROC curve (AUC) for logistic regression models fit on data representations extracted with a pre-trained diffusion model at ten different diffusion timesteps. High-grained features are already distinguishable at late diffusion steps (closer to random noise), while low-grained features are only represented at the earlier stage of the forward diffusion.}
%     \label{fig:celeba_attributes}
% \end{figure}
% % \end{wrapfigure}

%===SECTION===
\section{Method}
\label{sec:method}
% Taking into account the observations described in Section \ref{sec:diffusion_representations}, we propose to train a joint model that is composed of a classifier and a DDGM. Specifically, we propose to use a shared parameterization, namely, a shared encoder of the UNet architecture that serves as the generative part and for calculating pooled features for the classifier.

%===SubSECTION===
\subsection{Joint Diffusion Models: DDGMs with classifiers}

Taking into account the observations described in Section \ref{sec:diffusion_representations}, we propose to train a joint model that is composed of a classifier and a DDGM. We propose to use a shared parameterization, namely, a shared encoder of the UNet architecture that serves as the generative part and for calculating pooled features for the classifier. We pool the latent representations of the data from different levels of the UNet architecture into one vector $\rvz$. On top of this vector, we build a classifier model trained to assign a label to the data example represented by the vector $\rvz$.

In particular, we consider the following parameterization of a denoising diffusion model within a single diffusion timestep $t$, $p_\theta(\rvz_{t-1} | \rvz_t)$. We distinguish the encoder $e_\nu$ with parameters $\nu$ that maps input $\rvx_t$ into a set of vectors $ \mathcal{Z}_t= e_\nu(\rvx_t) $, where $\mathcal{Z}_t = \{\rvz_t^1, \rvz_t^2 \dots \rvz_t^n \} $, i.e., a set of representation vectors derived from each depth level of the UNet architecture. The second component of the denoising diffusion model is the decoder $d_\psi$ with parameters $\psi$ that reconstructs feature vectors into a denoised sample, $\rvx_{t-1} = d_\psi(\mathcal{Z}_t)$.
Together the encoder and the decoder form the denoising model $p_\theta$ with parameters $\theta = \{\nu,\psi\}$. Next, we introduce a third part of our model, which is the classifier $g_\omega$ with parameters $\omega$ that predicts target class $\hat{y} = g_\omega(\mathcal{Z}_t )$. The first layer of the classifier is the average pooling that results in a single representation $\rvz_t$. % where $z_t = pool(\{ z_t^1, z_t^2 \dots z_t^n \})$. \kamil{No idea how to write this pooling in a nice way}

\begin{figure}[h]
    \vskip -4mm
    \centering
    \begin{tabular}[t]{cc}
        \includegraphics[height=1.65cm]{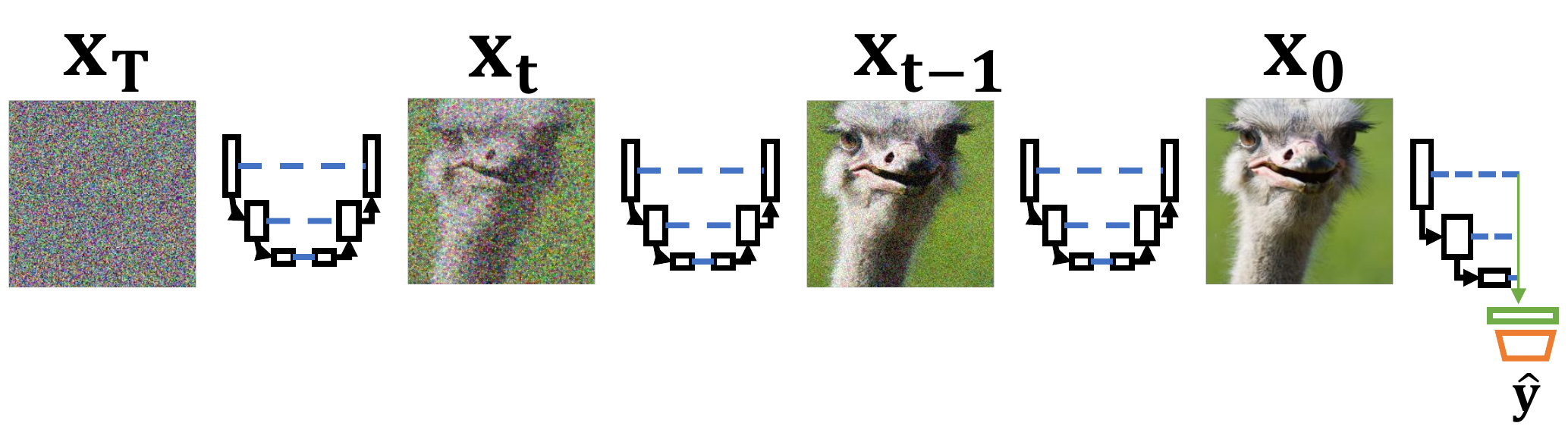} &
        % \hspace{1mm}\\&
        \includegraphics[height=1.65cm]{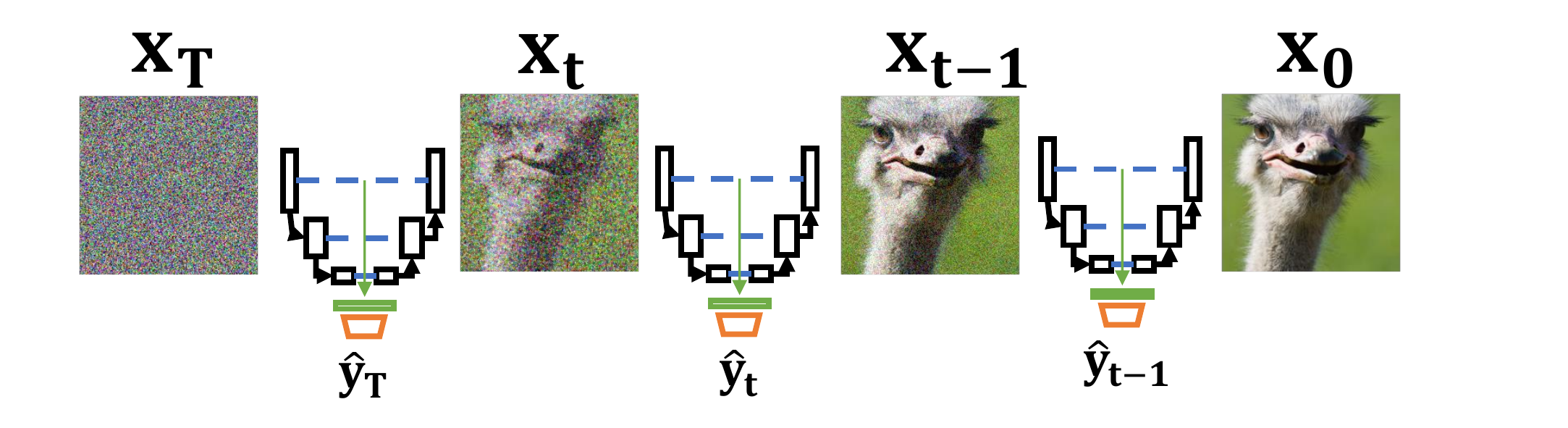} \\
        \textit{(a)} The parameterization of our joint diffusion  &\textit{(b)} Additional noisy classifiers\\
    \end{tabular}
    \vskip -1mm
    \caption{The parameterization of our joint diffusion model. \textit{(a)}~Each step in the backward diffusion is parameterized by a shared UNet. The classifier uses the encoder of the UNet together with the average pooling (green) and additional layers (yellow). \textit{(b)}~An alternative training that additionally uses the classifier for noisy images $\rvx_t$ ($t>0$).}
    \label{fig:different_joint_classifiers}
    \vskip  -6mm
\end{figure}

In our approach, we consider a classifier that takes the original image $\rvx_0$ for which a vector of probabilities is returned $\varphi$ and eventually the final prediction is calculated, $\hat{y} = g_{\omega}(\rvx_0)$. The visualization of our shared parameterization is presented in Fig.~\ref{fig:different_joint_classifiers}\textit{(a)}. As a result, our model could be written as follows $p_{\nu, \psi, \omega}(\rvx_{0:T}, y) = p_{\nu, \omega}(y | \rvx_0)\ p_{\nu, \psi}(\rvx_{0:T})$, 
% \begin{equation}\label{eq:joint}
%     p_{\nu, \psi, \omega}(\rvx_{0:T}, y) = p_{\nu, \omega}(y | \rvx_0)\ p_{\nu, \psi}(\rvx_{0:T}) ,
% \end{equation}
and applying the logarithm yields:
\begin{equation}\label{eq:joint_log}
    \ln p_{\nu, \psi, \omega}(\rvx_{0:T}, y) = \ln p_{\nu, \omega}(y | \rvx_0) + \ln p_{\nu, \psi}(\rvx_{0:T}).
\end{equation}
The logarithm of the joint distribution (\ref{eq:joint_log}) could serve as the training objective, where $\ln p_{\theta}(\rvx_{0:T})$ could be either approximated by the ELBO (\ref{eq:diff_elbo}) or the simplified objective with (\ref{eq:l_t_simple}). Here, we use the simplified objective:
\begin{equation}\label{eq:l_t_simple_ours}
    L_{t, \text{diff}}(\nu, \psi) = \mathbb{E}_{\mathbf{x}_{0}, \boldsymbol{\epsilon}}\left[\left\|\boldsymbol{\epsilon}-\boldsymbol{\hat{\epsilon}}\right\|^{2}\right],
\end{equation}
where $\hat{\epsilon}$ is a prediction from the decoder:
\begin{align}
     \{ \rvz_t^1, \rvz_t^2 \dots \rvz_t^n \} =& e_\nu\left(\sqrt{\overline{\alpha}_{t}} \mathbf{x}_{0}+\sqrt{1-\overline{\alpha}_{t}} \boldsymbol{\epsilon}, t\right) \\
     \boldsymbol{\hat{\epsilon}} =& d_\psi(\{ \rvz_t^1, \rvz_t^2 \dots \rvz_t^n \}) .
\end{align}
For the classifier, we use the logarithm of the categorical distribution:
\begin{equation}
% \resizebox{0.89\linewidth}{!}{$
    L_\text{class}(\nu, \omega) = -\E_{\mathbf{x}_0, y} \left[ \sum_{k=0}^{K-1} \mathds{1}[y=k] \log \frac{\exp \left( \varphi_{k} \right)}{\sum_{c=0}^{K-1} \exp \left( \varphi_{c} \right)} \right],
    % $}
\end{equation}
which is the cross-entropy loss, and where $y$ is the target class, $\varphi$ is a vector of probabilities returned by the classifier $g_{\omega}(e_{\nu}(\rvx_{0}))$, and $\mathds{1}[y=k]$ is the indicator function that is $1$ if $y$ equals $k$, and $0$ otherwise. 

The final loss function in our approach is then the following:
\begin{equation}\label{eq:final_joint}
    L(\nu, \psi, \omega) = L_\text{class}(\nu, \omega) - L_0(\nu, \psi) - \sum_{t=2}^{T} L_{t, \text{diff}}(\nu, \psi) - L_T(\nu, \psi) . \notag
\end{equation}
We optimize the objective in (\ref{eq:final_joint}) wrt. $\{ \nu, \psi, \omega \}$ with a single optimizer.
% We optimize the objective in (\ref{eq:final_joint}) jointly with a single optimizer over parameters $\{ \nu, \psi, \omega \}$.

%===SubSECTION===
\subsection{An alternative training of joint diffusion models}
\label{sect:alternative_training}

The training of the proposed approach over a batch of data is straightforward: For given $(\rvx_0, y)$, the example $\rvx_0$ is first noised with a forward diffusion to a random timestep, $\rvx_t$, so that the training loss for the denoising model is a Monte-Carlo approximation of the sum over all timesteps. Then $\rvx_0$ is fed to a classifier that returns probabilities $\varphi$, and the cross-entropy loss is calculated for given $y$.

As discussed in Section \ref{sect:diffusion_learn_features}, the diffusion model trained even in a fully unsupervised manner provides data representations related to the different granularity of input features at various diffusion timesteps. Thus, we can improve the robustness of our method by applying the same classifier to intermediate noisy images $\rvx_t$ ($0<t<T$), which by reason adds the cross-entropy losses for $\rvx_{t}$, namely: 
\begin{equation}
% \resizebox{0.89\linewidth}{!}{$
    L^t_\text{class}(\nu, \omega) = -\E_{\mathbf{x}_0, y} \left[ \sum_{k=0}^{K-1} \mathds{1}[y=k] \log \frac{\exp \left( \varphi_{k}^t \right)}{\sum_{c=0}^{K-1} \exp \left( \varphi_{c}^t \right)} \right],
    % $}
\end{equation}
where $\varphi_{k}^t$ is a vector of probabilities given by $g_{\omega}(e_{\nu}(\rvx_t))$. Then the extended objective (\ref{eq:final_joint}) is the following:
\begin{equation}\label{eq:alternative_objective}
    L_{\mathcal{T}}(\nu, \psi, \omega) = L(\nu, \psi, \omega) + \sum_{t \in \mathcal{T}} L^t_\text{class}(\nu, \omega) ,
\end{equation}
where $\mathcal{T} \subseteq \{1, 2, \ldots, T\}$ is the set of timesteps. These additional \textit{noisy classifiers} are schematically depicted in Fig.~\ref{fig:different_joint_classifiers}\textit{(b)} in which we highlight that the model is reused across various noisy images. It is important to mention that the noisy classifiers serve only for training purposes; they are not used for prediction. 
This procedure is similar to the data augmentation technique, where random noise is added to the input~\cite{sietsma1991creating}.
 
%===SubSECTION===
\subsection{Conditional sampling in joint diffusion models}
\label{sec:optimized_gen}
% \kamil{Classifier guidance as a name is already taken, although it'd fit the best. Any other idea?}
To improve the quality of samples generated by DDGM, \cite{dhariwal2021diffusion} propose a classifier guidance approach, where an externally trained classifier can be used to guide the generation of the DDGM trained in an unsupervised way towards the desired class. In DDGMs, at each backward diffusion step, an image is sampled from the output of the diffusion model $p_\theta$ according to the following formula:
\begin{equation}
    \begin{aligned}
    \mu, \Sigma \leftarrow& \mu_\theta\left(\rvx_t\right), \Sigma_\theta\left(\rvx_t\right)\\
    \rvx_{t-1} \leftarrow& \text{ sample from }\mathcal{N}\left(\mu,\Sigma\right)
\end{aligned}
\end{equation}
It was proposed in \cite{dhariwal2021diffusion} to change the second line of this equation and add a scaled gradient with respect to the target class from an externally trained classifier $c(\cdot)$ directly to the output of the denoising model:
\begin{equation}
    \begin{aligned}
    % \mu, \Sigma \leftarrow& \mu_\theta\left(x_t\right), \Sigma_\theta\left(x_t\right)\\
    \rvx_{t-1} \leftarrow& \text{ sample from }\mathcal{N}\left(\mu+s \Sigma \nabla_{\rvx_t} c(\rvx_t), \Sigma\right) ,
\end{aligned}
\end{equation}
where $s$ is a gradient scale.

With the joint training of a classifier and diffusion model introduced in this work, we propose to simplify the classifier guidance technique. 
Using the alternative training introduced in Section \ref{sect:alternative_training}, we can use noisy classifiers to formulate conditional sampling. The encoder model $e_\nu$ encodes input data $\rvx_t$ into the representation vectors $\mathcal{Z}_t $ that are used to both denoise an example into the previous diffusion timestep $\rvx_{t-1} \sim d_\psi\left(\mathcal{Z}_t \right)$ as well as to predict the target label with a classifier $\hat{y} = g_\omega\left(\mathcal{Z}_t \right)$. Therefore, to guide the model towards a target label during sampling, we propose optimizing the representations $\mathcal{Z}_t$ according to the gradient calculated through the classifier with respect to the desired class. The overview of this procedure is presented in Algorithm ~\ref{alg:guidance}.
\vskip -0.5cm

{\centering
\begin{minipage}{.7\linewidth}
\begin{algorithm}[H]
    \caption{Sampling with optimized representations given a diffusion model (an encoder $e_\nu(\mathcal{Z}_t|\rvx_t)$, a decoder $d_\phi(\rvx_{t-1}|\mathcal{Z}_t)$), a classifier $g_{\omega}(y|\mathcal{Z}_t)$, and a step size $\alpha$.}
    \label{alg:guidance}
    \begin{algorithmic}
        \STATE Input: class label $y$, step size $\alpha$
        \STATE $\rvx_T \gets \text{sample from } \mathcal{N}(0, \mathbf{I})$
        \FORALL{$t$ from $T$ to 1}
            \STATE $ \mathcal{Z}_t \leftarrow e_\nu(\rvx_t) $
            % \STATE $ \mathcal{L} \leftarrow 
            \STATE $ \mathcal{Z}_t'  \leftarrow \mathcal{Z}_t  - \alpha\nabla_{\mathcal{Z}_t}\log g_{\omega}(y| \mathcal{Z}_t )$
            \STATE $\mu, \Sigma \leftarrow d_\psi(\mathcal{Z}_t')$
            \STATE $\rvx_{t-1} \leftarrow \text{ sample from } \mathcal{N}(\mu , \Sigma)$
        \ENDFOR
        \STATE \textbf{return} $\rvx_0$
    \end{algorithmic}
    \end{algorithm}
\end{minipage}
}
\vskip 5mm

For the reformulation of the diffusion model proposed by~\cite{ho2020denoising} where instead of predicting the previous timestep $\rvx_{t-1}$ denoising model is optimized to predict noise $\epsilon$ that is subtracted from the image at the current timestep $\rvx_t$, we adequately change the optimization objective. Instead of optimizing the noise to be specific to the target class, we optimize it to be \textit{anything except for the target class}, which we implement by changing the optimization direction: $ \mathcal{Z}_t'  \leftarrow \mathcal{Z}_t  + \alpha\nabla_{\mathcal{Z}_t}\log g_{\omega}(y| \mathcal{Z}_t )$.

\section{Experiments}
\label{sec:experiments}

In the experiments, we aim for observing the benefits of the proposed joint diffusion model over a stand-alone classifier or a marginal diffusion model. To that end, we run a series of experiments to verify various properties, namely:
\begin{itemize}
    \item We measure the quality on a discriminative task, to evaluate whether training together with a diffusion model improves the robustness of the classifier.
    \item We measure the generative capability of our model to check if representations optimized by the classifier can lead to more accurate conditional generations.
    % \item We train our model in a semi-supervised setup to see if shared representations between the classifier and the diffusion model can positively influence the classification accuracy for a limited number of labeled data.
    % \item We use a domain-adaptation task to check if optimizing the representations using our approach helps to adapt to new data compared to a stand-alone classifier.
    \item We show that our joint model learns abstract features that can be used for the counterfactual explanation.
\end{itemize}

We use a UNet-based model with a depth level of three in all experiments. We pool its latent features with average pooling into a single vector, on top of which we add a classifier with two linear layers and the LeakyReLU activation. All metrics are reported for the standard training with the objective in (\ref{eq:final_joint}), except for the conditional sampling where we additionally train the classifier on noisy samples, i.e., additional losses as in (\ref{eq:alternative_objective}). Hyperparameters and training details are included in Appendix and code repository\footnote{\url{https://github.com/KamilDeja/joint_diffusion}}.

%===SubSECTION===
\subsection{Predictive performance of joint diffusion models}
In the first experiment, we evaluate the predictive performance of our method. To that end, we report the accuracy of our model on four datasets: FashionMNIST, SVHN, CIFAR-10, and CIFAR-100. We compare our method with a baseline classifier trained with a standard cross-entropy loss and the MLP classifier trained on top of representations extracted from the pre-trained DDGM as in Section~\ref{sec:diffusion_representations}, and three joint (hybrid) models: VERA \cite{grathwohl2021no}, JEM++ \cite{yang2021jem++}, HybViT \cite{yang2022your}. The results of this experiment are presented in Table \ref{tab:classifier_accuracy}.

\begin{table}[h!]
  \vskip -5mm
  \centering
  \caption{The classification accuracy calculated on the test sets. For each training of our methods and the vanilla classifier, we used exactly the same architectures.}
  \medskip
  % \resizebox{\linewidth}{!}{
  \def\arraystretch{1.25}
  \begin{tabular}{l|cccc}
    \hline
    Model & F-MNIST & SVHN & CIFAR-10 & CIFAR-100\\
    \midrule
    VERA \cite{grathwohl2021no} & - &  96.8\% & 93.2\% & 72.2\% \\
    JEM++ \cite{yang2021jem++} & -  & 96.9\% & 94.1\% & 74.5\% \\
    HybViT \cite{yang2022your} & -  & - & 95.9\% & 77.4\% \\
    \hline
    Classifier & 94.7\% & 96.9\% & 94.0\%  & 72.3\%\\
    % \hdashline 
    \makecell[l]{\textbf{Ours} (pre-trained DDGM)}& 60.6\%  & 79.6\% & 80.9\% & 45.9\%\\
    % \hdashline 
    \textbf{Ours} & \textbf{95.3\% }&\textbf{ 97.4\%} & \textbf{96.4\%} & \textbf{77.6\%} \\
    \hline
  \end{tabular}
  % }
  \label{tab:classifier_accuracy}
  % \vskip -5mm
\end{table}

As noticed before, a classifier trained on features extracted from the UNet of a DDGM pre-trained in an unsupervised manner achieves reasonable performance. However, it is always outperformed by a stand-alone classifier. The proposed joint diffusion model achieves the best performance on all four datasets. The reason for that could be two-fold. First, training a partially shared neural network (i.e., the encoder in the UNet architecture) benefits from the unsupervised training, similarly to how the pre-training using Boltzmann machines benefited finetuning of deep neural networks \cite{hinton2006fast}. Second, the shared encoder part is more robust since it is used in the backward diffusion for images with various levels of noise. 

\begin{table*}[htbp]
    \vskip -5mm
  \centering
  \caption{An evaluation of generative capabilities by measuring the FID score, Precision and Recall of generations from various diffusion-based models, including our joint diffusion model.}
  \medskip
  \resizebox{\linewidth}{!}{
  \def\arraystretch{1.25}
    \begin{tabular}{l||ccc|ccc|ccc|ccc}
        \hline
         \multirow{2}{*}{Model} & \multicolumn{3}{c|}{FashionMNIST} & \multicolumn{3}{c|}{CIFAR-10} & \multicolumn{3}{c|}{CIFAR-100}  & \multicolumn{3}{c}{CelebA} \\
        \cline{2-13}
         &  FID $\downarrow$&Prec $\uparrow$ & Rec $\uparrow$  & FID $\downarrow$ & Prec $\uparrow$ & Rec $\uparrow$ & FID $\downarrow$ & Prec $\uparrow$ & Rec $\uparrow$& FID $\downarrow$ & Prec $\uparrow$ & Rec $\uparrow$\\
        \midrule
        % VERA \cite{grathwohl2021no} & - & - & - & 27.5 & - & - & - & - & - & - & - & -  \\
        % JEM++ \cite{yang2021jem++} & - & - & - & 37.1 & - & - & - & - & - & - & - & -  \\
        % HybViT \cite{yang2022your} & - & - & - & 26.4 & - & - & 33.6 & - & - & - & - & -  \\
        % GenViT \cite{yang2022your} & - & - & - & 20.2 & - & - & 26.0 & - & - & 22.07 & - & -  \\
        % \midrule
        DDGM & 7.8 & \textbf{71.5} & \textbf{65.3} &  7.2 & 64.8 & 61.2 & 29.7 & \textbf{70.0}& 47.8& 5.6 &66.5 & \textbf{58.7}\\
        % \hdashline 
        \makecell[l]{DDGM (classifier guidance)} & 7.9 & 66.6& 59.5& 8.1 & 63.2 & \textbf{63.3} & 22.1 & 69.3 & 46.9 &4.9 & 66.0 & 57.8  \\
        % \hdashline 
        \midrule
        \textbf{Ours}& 8.7 & 71.1 & 61.1& 7.9 & 69.9 & 56.4  &17.4 & 63.2 & 54 & 7.0 & \textbf{67.5 }& 51.5  \\
        % \hdashline 
        \makecell[l]{\textbf{Ours} (conditional sampling)} & \textbf{5.9} & 63.1 & 63.2 & \textbf{6.4} & \textbf{70.7} & 54.3 & \textbf{16.8} & 63.5 & \textbf{54.1} &\textbf{4.8} &66.3 & 56.5 \\
    
        \hline
      \end{tabular}
  }
  \label{tab:fid_scores}
  \vskip -4mm
\end{table*}

\newpage
%===SubSECTION===
\subsection{Generative performance of joint diffusion models}

In the second experiment, we check how adding a classifier in our joint diffusion models influences the generative performance. We use the FID score to quantify the quality of data synthesis. Additionally, we use distributed Precision (Prec), and Recall (Rec) for assessing the exactness and diversity of generated samples~\cite{sajjadi2018assessing}. For our joint diffusion model, we consider samples from the prior let through the backward diffusion. We also use the second sampling scheme in which we use conditional sampling, namely, the optimization procedure as described in Section \ref{sec:optimized_gen}. 
We compare our approach with a vanilla DDGM, and a DDGM with classifier guidance \cite{dhariwal2021diffusion}, and recent state-of-the-art joint (hybrid) models: VERA \cite{grathwohl2021no}, JEM++ \cite{yang2021jem++}, HybViT and GenViT \cite{yang2022your}.

\begin{table*}[htbp]
    \vskip -9mm
  \centering
  \caption{A comparison of generative capabilities of joint models by measuring the FID score.}
  \medskip
  % \resizebox{\linewidth}{!}{
  \def\arraystretch{1.2}
    \begin{tabular}{l||c|c|c}
        \hline
         \multirow{2}{*}{Model}  & CIFAR-10 & CIFAR-100  & CelebA \\
        \cline{2-4}
         &  FID $\downarrow$ & FID $\downarrow$ &  FID $\downarrow$  \\
        \midrule
        VERA \cite{grathwohl2021no}  & 27.5 & - & -  \\
        JEM++ \cite{yang2021jem++}  & 37.1 & - & -  \\
        HybViT \cite{yang2022your}  & 26.4 & 33.6 & -   \\
        GenViT \cite{yang2022your}  & 20.2 & 26.0 & 22.07   \\
        \midrule
        \textbf{Ours} & 7.9  &17.4 &7.0   \\
        % \hdashline 
        \makecell[l]{\textbf{Ours} (conditional sampling)}  &  \textbf{6.4} & \textbf{16.8}  &\textbf{4.8}  \\
    
        \hline
      \end{tabular}
  % }
  \label{tab:fid_scores_sota}
  \vskip -5mm
\end{table*}

Overall, our proposition outperforms standard DDGMs regarding the general FID, see Table \ref{tab:fid_scores}. However, in some cases, the vanilla DDGM and the DDGM with the classifier guidance obtain better results in terms of the particular components: Precision (FashionMNIST, CIFAR-100) or Recall (FashionMNIST, CelebA). We can observe that conditional sampling improves the quality of generations in all evaluated benchmarks, especially in terms of precision that can be understood as the exactness of generations.
This could result from the fact that the optimization procedure drives $\mathcal{Z}_t$ to a mode. Eventually, the backward diffusion generates better samples. However, comparing our approach to current state-of-the-art joint models, we clearly outperform them all, see Table \ref{tab:fid_scores_sota}.

\begin{figure}[htbp!]
    % \vskip -3mm
    \centering
    \begin{tabular}{cc}
        \includegraphics[width=0.4\linewidth]{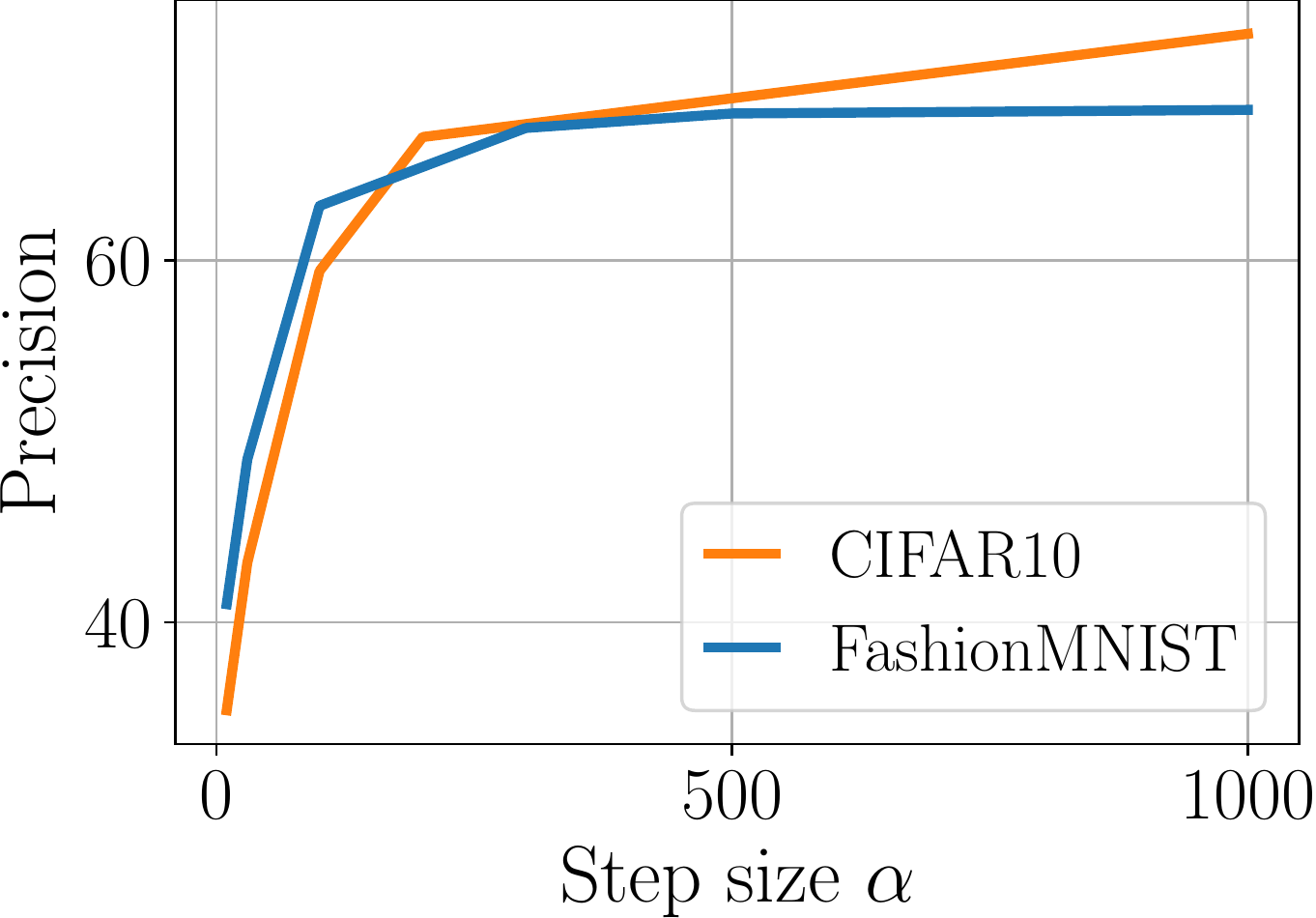} &
        \includegraphics[width=0.4\linewidth]{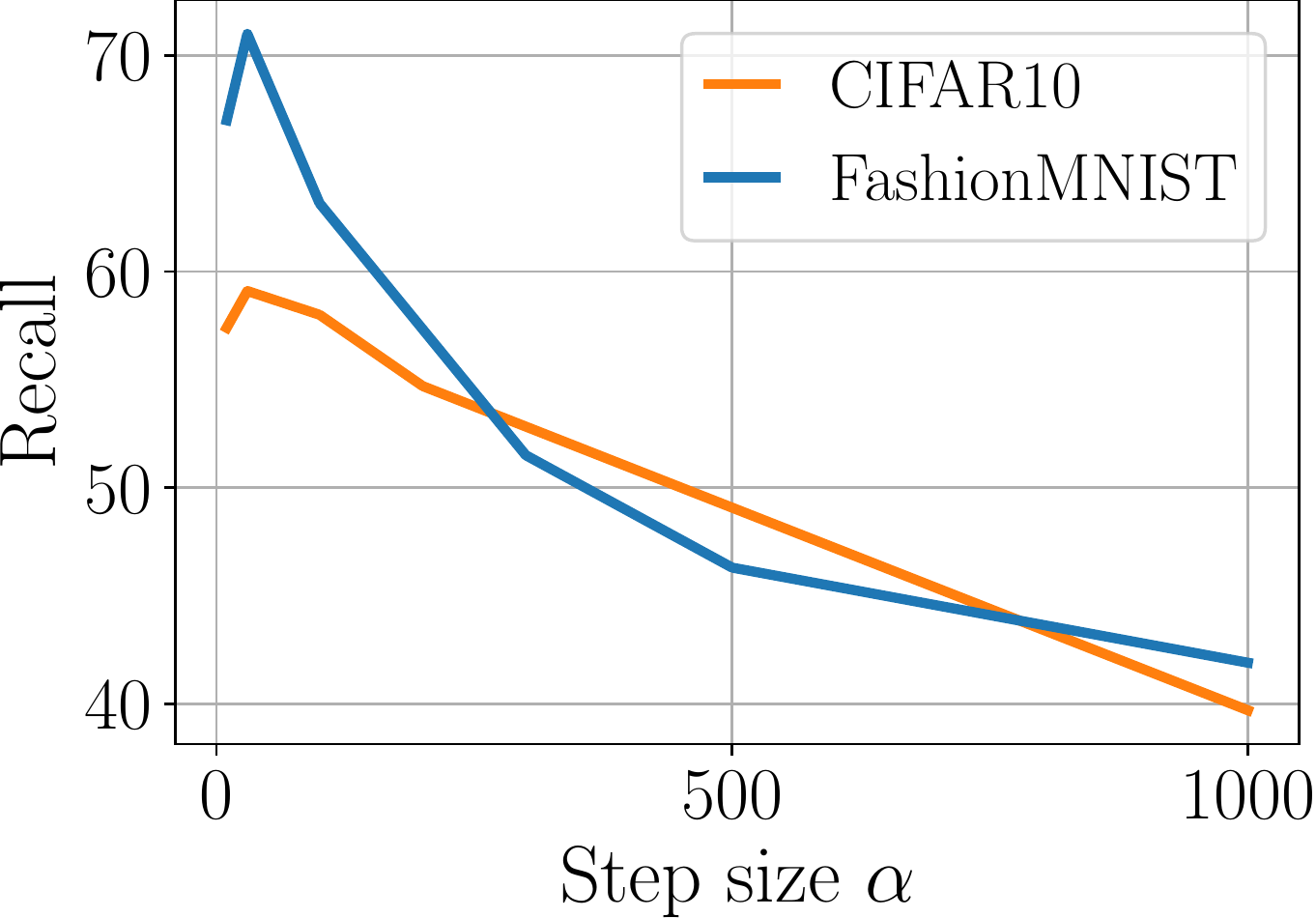} \\
        Precision  & Recall\\
    \end{tabular}
    \vskip -3mm
    \caption{The dependency between the value of the step size $\alpha$ and the value of Precision and Recall for the joint diffusion with conditional sampling.}
    \label{fig:optimization_prec_rec}
    % \vskip  -4mm
\end{figure}

\begin{figure}[htbp!]
    \vskip -3mm
    \centering
    \includegraphics[width=\linewidth]{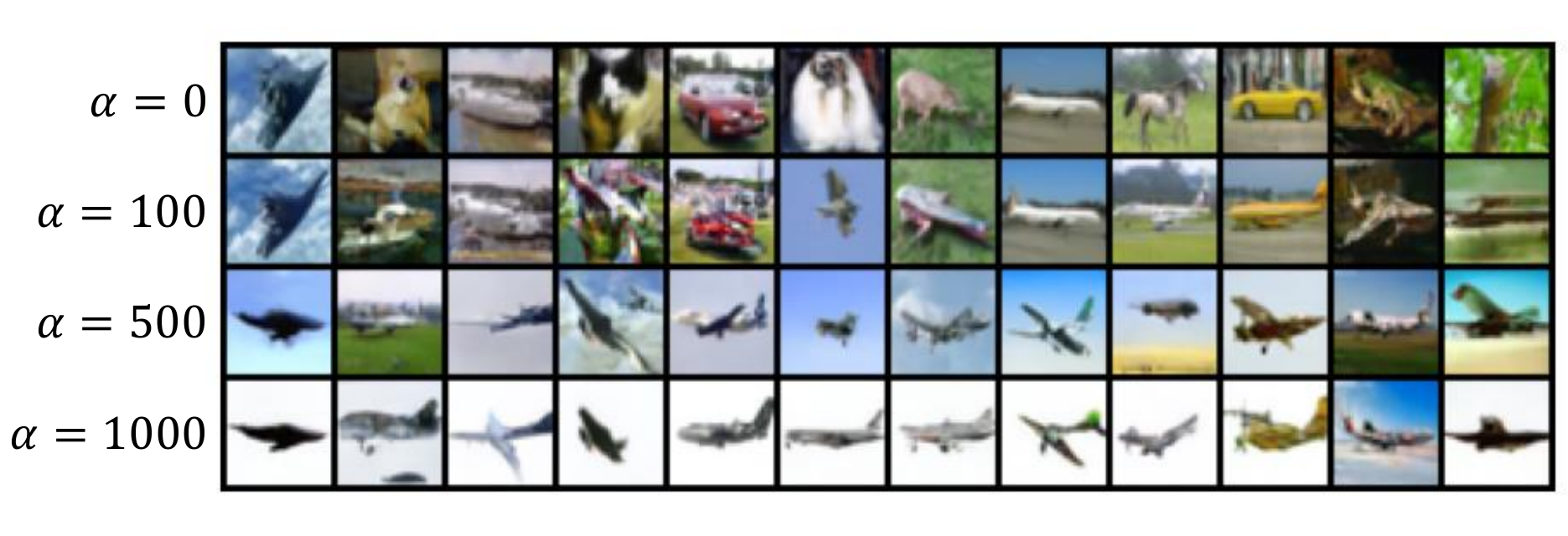}
    \vskip -3mm
    \caption{Samples from our joint diffusion model optimized towards a specific class (here: \emph{plane}) with different step sizes $\alpha$.}
    \label{fig:cifar_10_different_lr}
    \vskip -3mm
\end{figure}

To get further insight into the role of conditional sampling, we carried out an additional study for the varying value of $\alpha$ (the step size in Algorithm \ref{alg:guidance}). In Fig.~\ref{fig:optimization_prec_rec}, we present how Precision and Recall change for different values of this parameter. 
Apparently, increasing the step size value $\alpha$ leads to more precise but less diverse samples. This is rather intuitive behavior because larger steps result in features $\mathcal{Z}_t$ closer to modes. There seems to be a sweet spot around $\alpha \in [100, 250]$ for which both measures are high. Moreover, we visualize the effect of taking various values of $\alpha$ in Fig.~\ref{fig:cifar_10_different_lr}. For a chosen class, e.g., plane, we observe that the larger $\alpha$, the samples are more precise but they lack diversity (i.e., the background is almost the same).

\begin{figure}[!htbp]
    \vskip - 5mm
    \centering
    \begin{tabular}{cccc}
    \includegraphics[width=.25\linewidth]{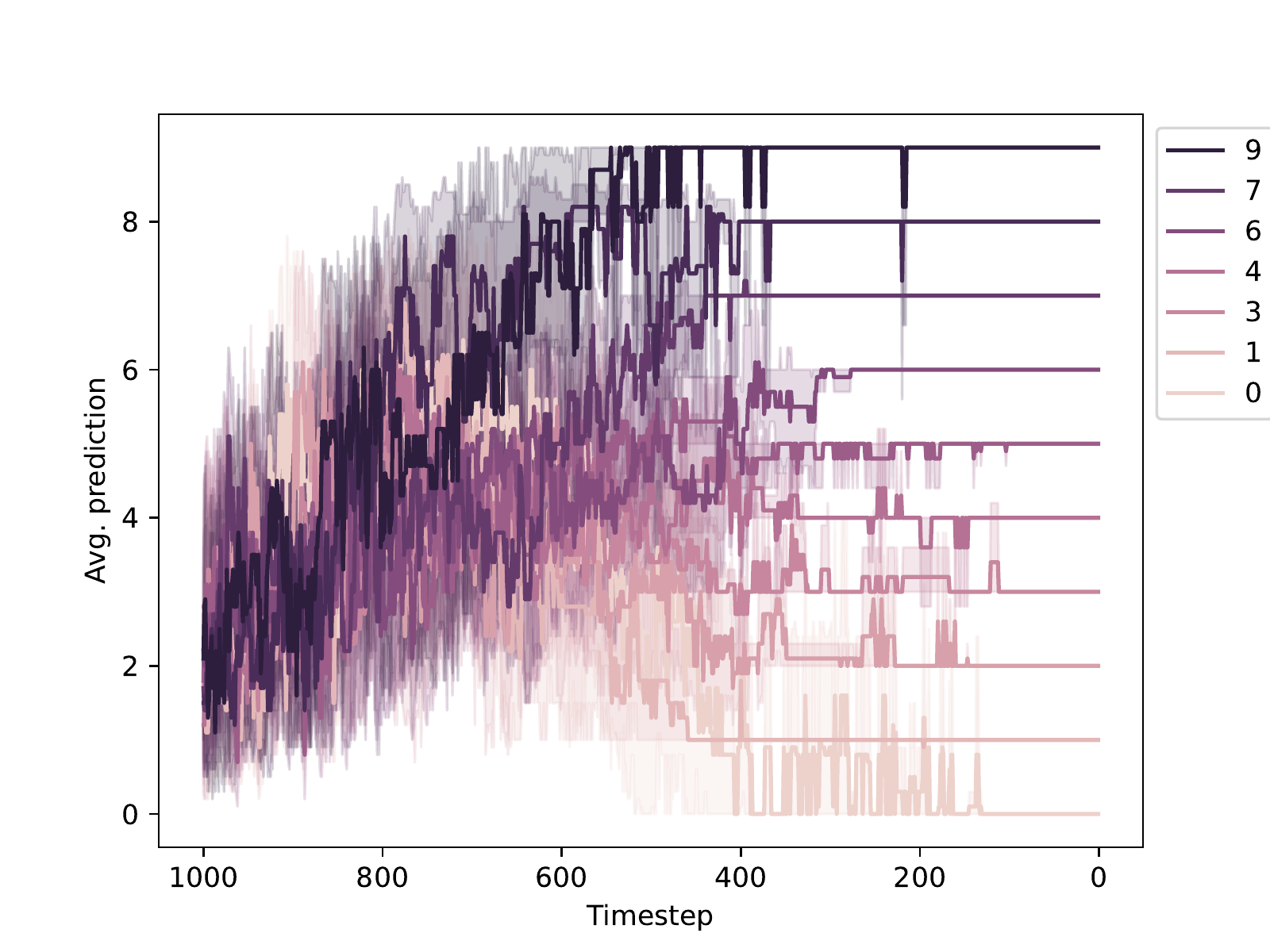} & \includegraphics[width=.23\linewidth]{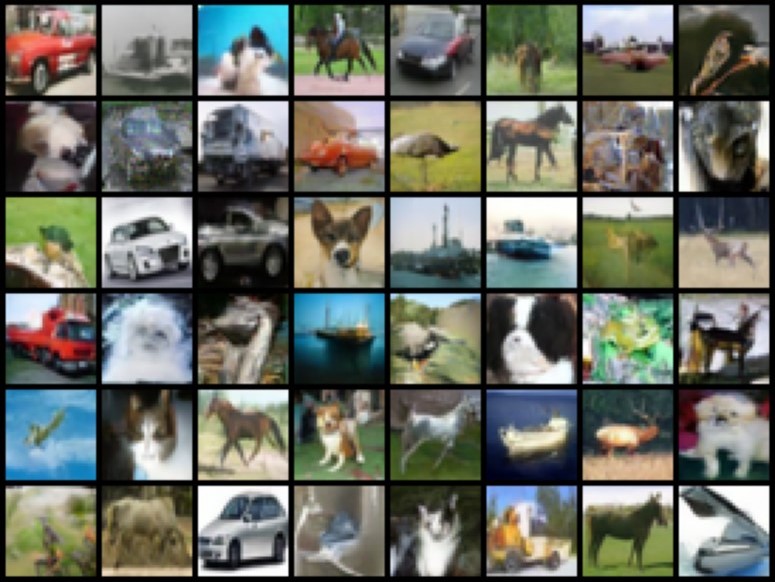} &
    \includegraphics[width=.25\linewidth]{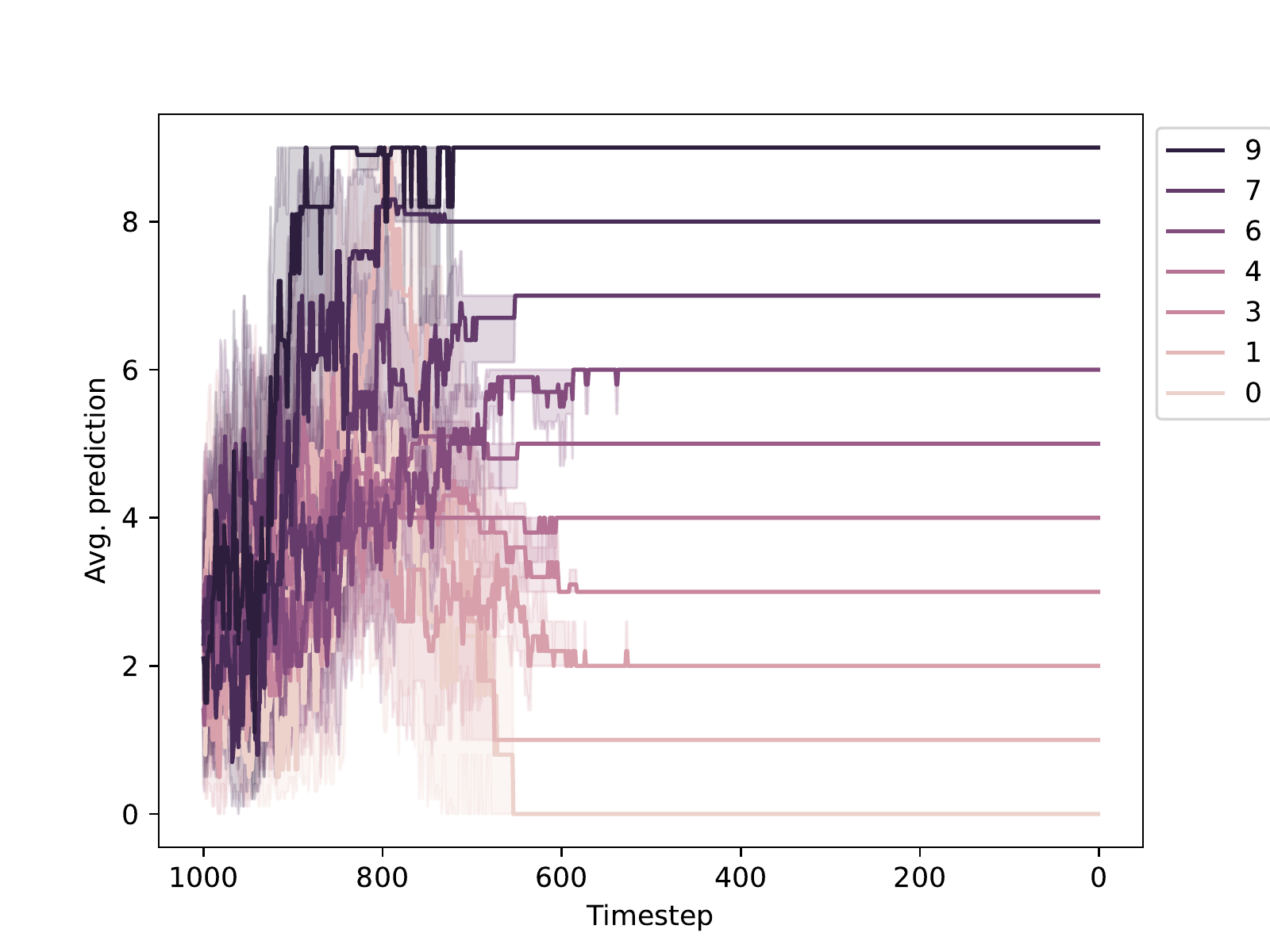} & \includegraphics[width=.23\linewidth]{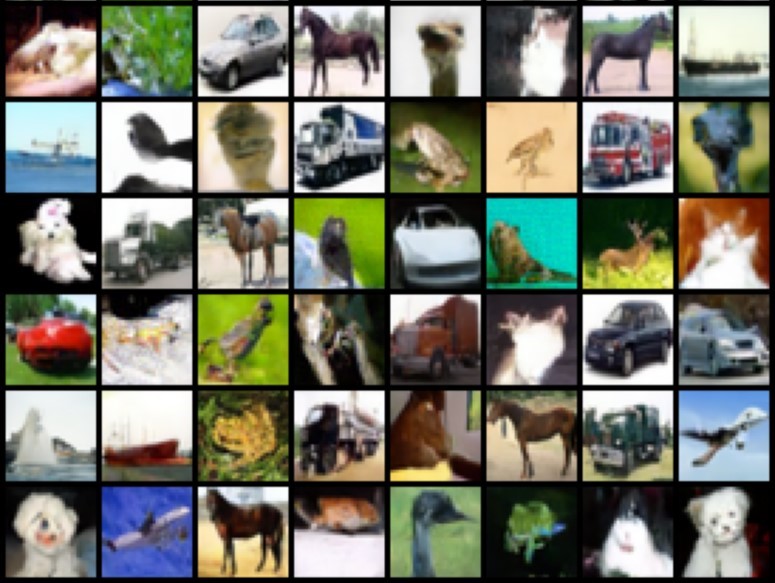}\\
    \multicolumn{2}{c}{\textit{(a)} $\alpha=200$} &\multicolumn{2}{c}{\textit{(b)} $\alpha=1000$}\
    \end{tabular}
    \vskip -2mm
    \caption{CIFAR10: Classifier decisions at different diffusion steps, for conditional sampling with different values of step size $\alpha$ and associated conditional samples}
    \vskip -5mm
    \label{fig:app_generations_decisions}
\end{figure}

In Fig.~\ref{fig:app_generations_decisions} we present how the decision of the classifier changes for sampling with the optimized generations. With a higher $\alpha$ step size value, optimization converges faster towards target classes. Interestingly, for the CIFAR10 dataset, there are certain classes (e.g., class 3) that converge later in the backward diffusion process than the others. 
We also present associated samples from our model. Once more, they depict that higher values of the $\alpha$ parameter lead to more precise but less diverse samples. We show more generations from our joint model in the Appendix~\ref{app:conditional_generation}.

 % \ref{app:conditional_generation}.

% \subsection{Interpolation in diffusion models' latent spaces}

%===SubSECTION===
\newpage
\subsection{A comparison to state-of-the-art joint approaches}
To get a better overview of the performance of our joint diffusion model, we present a comparison with other joint models and SOTA discriminative and generative models in Table \ref{tab:hybrid_results_comparison}. The purely discriminative and generative models are included as the upper bounds of the performance. Importantly, within the class of the joint models, our joint diffusion clearly outperforms all of the related works.

\begin{table}[htbp!]
    \vskip -8mm
    \centering
    \caption{A comparison of our joint diffusion model with other joint models, and the SOTA discriminative model, and the SOTA generative model on the CIFAR-10 test set.}
  \medskip
    % \resizebox{\linewidth}{!}{
    \def\arraystretch{1.25}
    \begin{tabular}{c|c | c c }
        \hline
        Class & Model & Accuracy\% $\uparrow$ & FID$\downarrow$ \\
        \hline
        \multirow{9}{*}{\textbf{Joint}} &  IGEBM \cite{du2019implicit} & 49.1 & 37.9\\
         & Glow \cite{kingma2018glow} &  67.6  &  48.9\\
         & Residual Flows \cite{chen2019residual} & 70.3 &  46.4\\
         & JEAT \cite{grathwohl2019classifier} & $85.2$  & 38.2 \\
        % & JEM $p(x|y)$ factored & 30.1 &  61.8 \\
         & JEM \cite{grathwohl2019classifier} & $92.9$  & 38.4 \\
         & VERA ($\alpha=100$) \cite{grathwohl2021no} & $93.2$  & 30.5 \\
         & JEM++ \cite{yang2021jem++} & $94.1$  & 38.0 \\
         & HybViT \cite{yang2022your} & $95.9$  & 26.4 \\
         & \textbf{Ours} & $\mathbf{96.4}$ & $\mathbf{7.9}$\\
        \hline
        % \hline
        \multirow{1}{*}{\textbf{Disc.}} & VIT-H \cite{dosovitskiy2020image} & 99.5 & - \\
        \hline
        % \multirow{2}{*}{\textbf{Gen.}} & NCSN & - & 25.32 \\
        \multirow{2}{*}{\textbf{Gen.}}& DDGM (our implementation) & - &7.2 \\
         & LSGM \cite{vahdat2021score} & - & 2.1 \\
        
        \hline
    \end{tabular}
    % }
    \label{tab:hybrid_results_comparison}
    \vskip -5mm
\end{table}

\subsection{Visual Counterfactual Explanations}
\label{sec:medical}

% \begin{figure}[htbp!]
%     \vskip -2.5mm
%     \centering
%     \begin{tabular}{cc}
%         \includegraphics[width=0.375\linewidth]{Figures/malaria_negative.png} &
%         \includegraphics[width=0.375\linewidth]{Figures/malaria_positive.png} \\
%         Negative examples & Positive examples\\
%     \end{tabular}
%     \vskip -2.5mm
    
%     \caption{Data samples from the Malaria dataset classified as negative examples (left) or parasitized cells (right). (\textit{top row}) original data examples, ($2^{nd}$ \textit{row}) data noised with 20\% of forward diffusion steps, ($3^{rd}$ \textit{row}) denoised images with conditional sampling, (\textit{bottom row}) the difference between the $3^{rd}$ and $4^{th}$ rows.}
%     \label{fig:malaria}
%     \vskip  -4mm
% \end{figure}

In the last experiment, we apply our joint diffusion model to real-world medical data, the MALARIA dataset~\cite{rajaraman2018pre}, that includes 27,558 cell images that are either infected by the malaria parasite or not (a classification task). The cells have various shapes and different staining (i.e., colors) and contain or not the parasite (visually apparent as a purple dot). 

% \begin{wrapfigure}{r}{0.6\textwidth}
%     \vskip -10mm
%     \begin{adjustbox}{center}
%     \begin{tabular}{cc}
%         \includegraphics[width=0.375\linewidth]{Figures/malaria_negative.png} &
%         \includegraphics[width=0.375\linewidth]{Figures/malaria_positive.png} \\
%         Negative examples & Positive examples\\
%     \end{tabular}
%     \end{adjustbox}
%     \vskip -2mm
%     \caption{Data samples from the Malaria dataset classified as negative examples (left) or parasitized cells (right). (\textit{top row}) original data examples, ($2^{nd}$ \textit{row}) data noised with 20\% of forward diffusion steps, ($3^{rd}$ \textit{row}) denoised images with conditional sampling, (\textit{bottom row}) the difference between the $3^{rd}$ and $4^{th}$ rows. model.}
%     \label{fig:malaria}
%     \vskip -5mm
% \end{wrapfigure}

\begin{figure}[htbp!]
    \vskip -2.5mm
    \centering
    \begin{tabular}{cc}
        \includegraphics[width=0.4\linewidth]{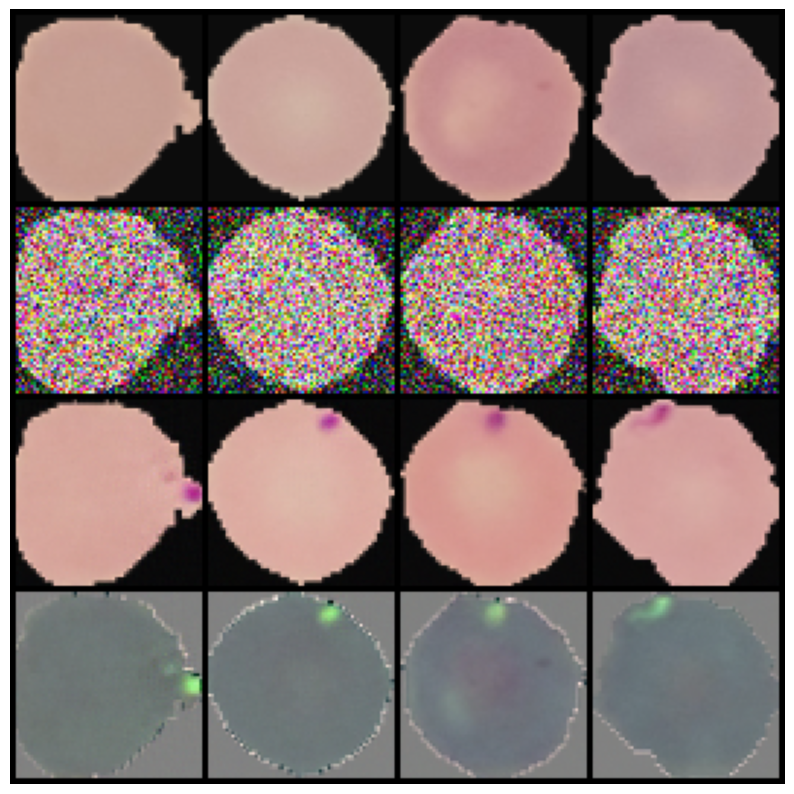} &
        \includegraphics[width=0.4\linewidth]{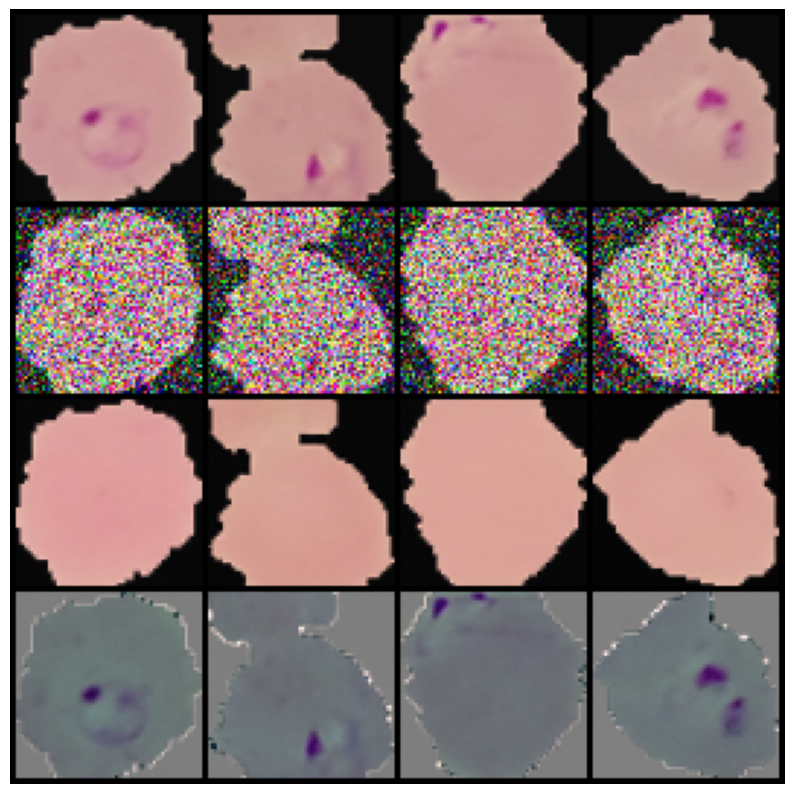} \\
        Negative examples & Positive examples\\
    \end{tabular}
    \vskip -2.5mm
    
    \caption{Data samples from the Malaria dataset classified as negative examples (left) or parasitized cells (right). (\textit{top row}) original data examples, ($2^{nd}$ \textit{row}) data noised with 20\% of forward diffusion steps, ($3^{rd}$ \textit{row}) denoised images with conditional sampling, (\textit{bottom row}) the difference between the $3^{rd}$ and $4^{th}$ rows.}
    \label{fig:malaria}
    \vskip  -4mm
\end{figure}

After training our joint diffusion model, we obtain high classification accuracy ($98\%$) on the test set. On top of this, we introduce an adaptation of visual counterfactual explanations (VCE) method~\cite{augustin2022diffusion} that provides an answer to the question: \textit{What is the minimal change to the input image $\rvx_0$ to change the decision of the classifier}. In our setup, we answer this question with a conditional sampling algorithm that we use to generate the counterfactual explanations.

In Fig.~\ref{fig:malaria}, we show a few examples from the negative (left) or positive (right) classes. 
We add $20\%$ of noise to these images and run conditional sampling with the opposite class (i.e., changing negative examples to positive ones and \textit{vice versa}). In both cases, the joint diffusion model with conditional sampling can either remove the parasite from the positive examples or add the parasite to the negative ones. All presented images are not cherry-picked.

This experiment shows that not only we can use our proposed approach to obtain a powerful classifier but also to visualize its regions of interest. In the considered case, calculating the difference between the original example and the image with a changed class label indicates the malaria plasmodium (see the last row in Fig.~\ref{fig:malaria}). We provide more examples from the CelebA data in Appendix~\ref{app:counterfactual_generation}.  

%===SECTION===
\section{Conclusion}

In this work, we introduced a joint model that combines a diffusion model and a classifier through shared parameterization. We first experimentally demonstrated that DDGMs learn semantically meaningful data representations that could be used for classification. On top of this observation, we introduced our joint diffusion models.
In the experimental section, we showed that our approach improves the performance in both the classification and generative tasks, providing high-quality generations and enabling conditional generations with built-in classifier guidance. Our proposed approach achieves state-of-the-art performance in the class of joint models. Additionally, we show that the joint diffusion model can be used for visual counterfactual explanations without any changes to the original setup.
\newpage
\bibliographystyle{splncs04}
\bibliography{bibliography}

\begin{thebibliography}{10}
\providecommand{\url}[1]{\texttt{#1}}
\providecommand{\urlprefix}{URL }
\providecommand{\doi}[1]{https://doi.org/#1}

\bibitem{abstreiter2021diffusionbased}
Abstreiter, K., Mittal, S., Bauer, S., Schölkopf, B., Mehrjou, A.:
  Diffusion-based representation learning. arXiv preprint arXiv:
  Arxiv-2105.14257  (2021)

\bibitem{augustin2022diffusion}
Augustin, M., Boreiko, V., Croce, F., Hein, M.: Diffusion visual counterfactual
  explanations. arXiv preprint arXiv:2210.11841  (2022)

\bibitem{baranchuk2021labelefficient}
Baranchuk, D., Rubachev, I., Voynov, A., Khrulkov, V., Babenko, A.:
  Label-efficient semantic segmentation with diffusion models. International
  Conference On Learning Representations  (2021)

\bibitem{chandra2014adaptive}
Chandra, B., Sharma, R.K.: Adaptive noise schedule for denoising autoencoder.
  In: International conference on neural information processing. pp. 535--542.
  Springer (2014)

\bibitem{chapelle2009semi}
Chapelle, O., Scholkopf, B., Zien, A.: Semi-supervised learning (chapelle, o.
  et al., eds.; 2006)[book reviews]. IEEE Transactions on Neural Networks
  \textbf{20}(3),  542--542 (2009)

\bibitem{chen2019residual}
Chen, R.T., Behrmann, J., Duvenaud, D., Jacobsen, J.H.: Residual flows for
  invertible generative modeling. arXiv preprint arXiv:1906.02735  (2019)

\bibitem{dhariwal2021diffusion}
Dhariwal, P., Nichol, A.: Diffusion models beat {GANs} on image synthesis.
  Advances in Neural Information Processing Systems  \textbf{34} (2021)

\bibitem{dosovitskiy2020image}
Dosovitskiy, A., Beyer, L., Kolesnikov, A., Weissenborn, D., Zhai, X.,
  Unterthiner, T., Dehghani, M., Minderer, M., Heigold, G., Gelly, S.,
  Uszkoreit, J., Houlsby, N.: An image is worth 16x16 words: Transformers for
  image recognition at scale. International Conference On Learning
  Representations  (2020)

\bibitem{du2019implicit}
Du, Y., Mordatch, I.: Implicit generation and generalization in energy-based
  models. arXiv preprint arXiv:1903.08689  (2019)

\bibitem{esser2018variational}
Esser, P., Sutter, E., Ommer, B.: A variational u-net for conditional
  appearance and shape generation. Ieee/cvf Conference On Computer Vision And
  Pattern Recognition  (2018). \doi{10.1109/CVPR.2018.00923}

\bibitem{falck2022a}
Falck, F., Williams, C., Danks, D., Deligiannidis, G., Yau, C., Holmes, C.C.,
  Doucet, A., Willetts, M.: A multi-resolution framework for u-nets with
  applications to hierarchical {VAE}s. In: Oh, A.H., Agarwal, A., Belgrave, D.,
  Cho, K. (eds.) Advances in Neural Information Processing Systems (2022)

\bibitem{ganin2016domain}
Ganin, Y., Ustinova, E., Ajakan, H., Germain, P., Larochelle, H., Laviolette,
  F., Marchand, M., Lempitsky, V.: Domain-adversarial training of neural
  networks. The journal of machine learning research  \textbf{17}(1),
  2096--2030 (2016)

\bibitem{geras2014scheduled}
Geras, K.J., Sutton, C.: Scheduled denoising autoencoders. arXiv preprint
  arXiv:1406.3269  (2014)

\bibitem{grathwohl2019classifier}
Grathwohl, W., Wang, K.C., Jacobsen, J., Duvenaud, D., Norouzi, M., Swersky,
  K.: Your classifier is secretly an energy based model and you should treat it
  like one. International Conference On Learning Representations  (2019)

\bibitem{grathwohl2019your}
Grathwohl, W., Wang, K.C., Jacobsen, J.H., Duvenaud, D., Norouzi, M., Swersky,
  K.: Your classifier is secretly an energy based model and you should treat it
  like one. In: International Conference on Learning Representations (2019)

\bibitem{grathwohl2021no}
Grathwohl, W.S., Kelly, J.J., Hashemi, M., Norouzi, M., Swersky, K., Duvenaud,
  D.: No {\{}mcmc{\}} for me: Amortized sampling for fast and stable training
  of energy-based models. In: International Conference on Learning
  Representations (2021)

\bibitem{hinton2006fast}
Hinton, G.E., Osindero, S., Teh, Y.W.: A fast learning algorithm for deep
  belief nets. Neural computation  \textbf{18}(7),  1527--1554 (2006)

\bibitem{ho2020denoising}
Ho, J., Jain, A., Abbeel, P.: Denoising diffusion probabilistic models.
  Advances in Neural Information Processing Systems  \textbf{33},  6840--6851
  (2020)

\bibitem{ho2022classifierfree}
Ho, J., Salimans, T.: Classifier-free diffusion guidance. arXiv preprint arXiv:
  Arxiv-2207.12598  (2022)

\bibitem{huang2021variational}
Huang, C.W., Lim, J.H., Courville, A.C.: A variational perspective on
  diffusion-based generative models and score matching. Advances in Neural
  Information Processing Systems  \textbf{34} (2021)

\bibitem{huang2022improving}
Huang, P.K.M., Chen, S.A., Lin, H.T.: Improving conditional score-based
  generation with calibrated classification and joint training. In: NeurIPS
  2022 Workshop on Score-Based Methods

\bibitem{ilse2020diva}
Ilse, M., Tomczak, J.M., Louizos, C., Welling, M.: Diva: Domain invariant
  variational autoencoders. In: Medical Imaging with Deep Learning. pp.
  322--348. PMLR (2020)

\bibitem{jebara2012machine}
Jebara, T.: Machine learning: discriminative and generative, vol.~755. Springer
  Science \& Business Media (2012)

\bibitem{jin2017introspective}
Jin, L., Lazarow, J., Tu, Z.: Introspective classification with convolutional
  nets. Advances in Neural Information Processing Systems  \textbf{30} (2017)

\bibitem{kingma2021variational}
Kingma, D.P., Salimans, T., Poole, B., Ho, J.: Variational diffusion models.
  In: Advances in Neural Information Processing Systems (2021)

\bibitem{kingma2014autoencoding}
Kingma, D.P., Welling, M.: Auto-{E}ncoding {V}ariational {B}ayes. In: ICLR
  (2014)

\bibitem{kingma2018glow}
Kingma, D.P., Dhariwal, P.: Glow: Generative flow with invertible 1x1
  convolutions. In: Advances in Neural Information Processing Systems. pp.
  10215--10224 (2018)

\bibitem{kingma2014semi}
Kingma, D.P., Mohamed, S., Jimenez~Rezende, D., Welling, M.: Semi-supervised
  learning with deep generative models. Advances in neural information
  processing systems  \textbf{27} (2014)

\bibitem{knop2020cramer}
Knop, S., Spurek, P., Tabor, J., Podolak, I., Mazur, M., Jastrzebski, S.:
  Cramer-wold auto-encoder. The Journal of Machine Learning Research
  \textbf{21}(1),  6594--6621 (2020)

\bibitem{lasserre2006principled}
Lasserre, J.A., Bishop, C.M., Minka, T.P.: Principled hybrids of generative and
  discriminative models. In: 2006 IEEE Computer Society Conference on Computer
  Vision and Pattern Recognition (CVPR'06). vol.~1, pp. 87--94. IEEE (2006)

\bibitem{lazarow2017introspective}
Lazarow, J., Jin, L., Tu, Z.: Introspective neural networks for generative
  modeling. In: Proceedings of the IEEE International Conference on Computer
  Vision. pp. 2774--2783 (2017)

\bibitem{lee2018wasserstein}
Lee, K., Xu, W., Fan, F., Tu, Z.: Wasserstein introspective neural networks.
  In: Proceedings of the IEEE Conference on Computer Vision and Pattern
  Recognition. pp. 3702--3711 (2018)

\bibitem{masarczyk2021onrob}
Masarczyk, W., Deja, K., Trzcinski, T.: On robustness of generative
  representations against catastrophic forgetting. In: Mantoro, T., Lee, M.,
  Ayu, M.A., Wong, K.W., Hidayanto, A.N. (eds.) Neural Information Processing.
  pp. 325--333. Springer International Publishing, Cham (2021)

\bibitem{nalisnick2019hybrid}
Nalisnick, E., Matsukawa, A., Teh, Y.W., Gorur, D., Lakshminarayanan, B.:
  Hybrid models with deep and invertible features. In: International Conference
  on Machine Learning. pp. 4723--4732. PMLR (2019)

\bibitem{nichol2021improved}
Nichol, A.Q., Dhariwal, P.: Improved denoising diffusion probabilistic models.
  In: International Conference on Machine Learning. pp. 8162--8171. PMLR (2021)

\bibitem{perugachi2021invertible}
Perugachi-Diaz, Y., Tomczak, J., Bhulai, S.: Invertible densenets with
  concatenated lipswish. Advances in Neural Information Processing Systems
  \textbf{34},  17246--17257 (2021)

\bibitem{rajaraman2018pre}
Rajaraman, S., Antani, S.K., Poostchi, M., Silamut, K., Hossain, M.A., Maude,
  R.J., Jaeger, S., Thoma, G.R.: Pre-trained convolutional neural networks as
  feature extractors toward improved malaria parasite detection in thin blood
  smear images. PeerJ  \textbf{6},  e4568 (2018)

\bibitem{ronneberger2015u}
Ronneberger, O., Fischer, P., Brox, T.: U-net: Convolutional networks for
  biomedical image segmentation. In: International Conference on Medical image
  computing and computer-assisted intervention. pp. 234--241. Springer (2015)

\bibitem{sajjadi2018assessing}
Sajjadi, M.S., Bachem, O., Lucic, M., Bousquet, O., Gelly, S.: Assessing
  generative models via precision and recall. arXiv preprint arXiv:1806.00035
  (2018)

\bibitem{sietsma1991creating}
Sietsma, J., Dow, R.J.: Creating artificial neural networks that generalize.
  Neural networks  \textbf{4}(1),  67--79 (1991)

\bibitem{sohl2015deep}
Sohl-Dickstein, J., Weiss, E., Maheswaranathan, N., Ganguli, S.: Deep
  unsupervised learning using nonequilibrium thermodynamics. In: International
  Conference on Machine Learning. pp. 2256--2265. PMLR (2015)

\bibitem{song2019generative}
Song, Y., Ermon, S.: Generative modeling by estimating gradients of the data
  distribution. Advances in Neural Information Processing Systems  \textbf{32}
  (2019)

\bibitem{song2020score}
Song, Y., Sohl-Dickstein, J., Kingma, D.P., Kumar, A., Ermon, S., Poole, B.:
  Score-based generative modeling through stochastic differential equations.
  In: International Conference on Learning Representations (2020)

\bibitem{NEURIPS2021_cfe8504b}
Tashiro, Y., Song, J., Song, Y., Ermon, S.: Csdi: Conditional score-based
  diffusion models for probabilistic time series imputation. In: Advances in
  Neural Information Processing Systems. vol.~34, pp. 24804--24816. Curran
  Associates, Inc. (2021)

\bibitem{tomczak2022deep}
Tomczak, J.M.: Deep Generative Modeling. Springer Cham (2022)

\bibitem{tulyakov2017hybrid}
Tulyakov, S., Fitzgibbon, A., Nowozin, S.: Hybrid vae: Improving deep
  generative models using partial observations. arXiv preprint arXiv:1711.11566
   (2017)

\bibitem{tzen2019neural}
Tzen, B., Raginsky, M.: Neural stochastic differential equations: Deep latent
  gaussian models in the diffusion limit. arXiv preprint arXiv:1905.09883
  (2019)

\bibitem{vahdat2021score}
Vahdat, A., Kreis, K., Kautz, J.: Score-based generative modeling in latent
  space. Advances in Neural Information Processing Systems  \textbf{34} (2021)

\bibitem{yang2022chroma}
Yang, W., Kirichenko, P., Goldblum, M., Wilson, A.G.: Chroma-vae: Mitigating
  shortcut learning with generative classifiers. arXiv preprint
  arXiv:2211.15231  (2022)

\bibitem{yang2021jem++}
Yang, X., Ji, S.: Jem++: Improved techniques for training jem. In: Proceedings
  of the IEEE/CVF International Conference on Computer Vision. pp. 6494--6503
  (2021)

\bibitem{yang2022your}
Yang, X., Shih, S.M., Fu, Y., Zhao, X., Ji, S.: Your vit is secretly a hybrid
  discriminative-generative diffusion model. arXiv preprint arXiv:2208.07791
  (2022)

\bibitem{zhang2018convolutional}
Zhang, Q., Zhang, L.: Convolutional adaptive denoising autoencoders for
  hierarchical feature extraction. Frontiers of Computer Science
  \textbf{12}(6),  1140--1148 (2018)

\end{thebibliography}

\newpage
\appendix
\section*{Appendix}

\section{Additional experiments}
With state-of-the-art performance in joint modeling, we further evaluate other setups where one part of our model can benefit from another. In particular, we propose two additional proof-of-concept experiments (with smaller architectures):

\begin{itemize}
    \item We train our model in a semi-supervised setup to see if shared representations between the classifier and the diffusion model can positively influence the classification accuracy for a limited number of labeled data.
    \item We use a domain-adaptation task to check if optimizing the representations using our approach helps to adapt to new data compared to a stand-alone classifier.
\end{itemize}

\subsection{Semi-supervised learning of joint diffusion models}
% \textbf{Can one part of the model benefit from representations optimized with another?}
We evaluate our approach in the semi-supervised setup, where we artificially limit the amount of labeled data to $10\%$, $5\%$ or $1\%$ in three datasets SVHN, CIFAR-10, and CIFAR-100.
We compare joint diffusion models to a deep neural network-based classifier and a deep neural network-based classifier on top of the pre-trained UNet encoder. The results are presented in Table~\ref{tab:acc_semi}. For simplicity, in this setup, we do not use \emph{any} data augmentation technique, therefore the performance on the full dataset is slightly lower than what we presented in Table~\ref{tab:classifier_accuracy}.

In the case of the stand-alone classifier, we observe that classification accuracy drastically drops with the number of labeled data. However, in our joint diffusion model, we can train the classifier on the smaller dataset while still optimizing the generator part in an unsupervised manner, with all available unlabelled data. This approach significantly improves the classifier's performance thanks to the improved quality of data representations. For CIFAR-10, we observe that the joint diffusion model with only 5\% of labeled data (250 examples per class) performs almost as well as the stand-alone classifier trained with the fully labeled training dataset. In more extreme scenarios, e.g., labeled data limited to 50, 25, or 5 examples per class, it seems to be slightly more beneficial to first learn the data representation in an unsupervised way and then add the classifier on top of them. However, overall, the joint diffusion model performs extremely well and greatly benefits from available unlabeled data in terms of classification accuracy. Our experiments align with the observation by~\cite{baranchuk2021labelefficient}, where DDGMs were used to improve the performance in semi-supervised image segmentation. 

\begin{table*}[h!tbp]
  \vskip -4mm
  \centering
  \caption{The accuracy of the classifier trained in the semi-supervised setup, for each dataset we train the classifier with the fully labeled data or a limited amount of labeled examples and the remaining unlabelled examples. We compare the standard classifier with the classifier trained on a pre-trained DDGM as presented in Sec~\ref{sec:diffusion_representations} and our joint diffusion method.}
  \smallskip
  \resizebox{\linewidth}{!}{
    \def\arraystretch{1.1}
    \begin{tabular}{l||ccc|ccc|cccc}
        \hline
         & \multicolumn{3}{c|}{SVHN} & \multicolumn{3}{c|}{CIFAR-10} &   \multicolumn{4}{c}{CIFAR-100} \\
        \hline
         Labelled data&  100\% & 5\%& 1\% & 100\% & 5\%& 1\%& 100\% & 10\% & 5\%& 1\%\\
         Images per class &10000 & 500 & 100 & 5000 & 250 & 50 & 500 & 50 & 25 & 5\\
        \midrule
        Classifier & 95.1 & 87.8 &75.15 & 81 &46.4 & 31.5 & 60.8 & 22.2 & 16.6 & 6.9 \\
        % \hdashline 
        \makecell[l]{\textbf{Ours} (pre-trained DDGM)}& 79.6 & 51.7 & 66.0 & 80.9 & 75.1 &\textbf{65.3 }& 43.8 & 33.9 & \textbf{28.8} & \textbf{15.4}  \\
        % \hdashline 
        \textbf{Ours} & \textbf{95.4} & \textbf{90.2} & \textbf{76.7} & \textbf{89.9} &\textbf{78.2} & 64.7 & \textbf{63.6} & \textbf{38.6} & 21.5 & 11.5 \\
    
        \hline
      \end{tabular}
  }
  \label{tab:acc_semi}
  % \vskip -5mm
\end{table*}

%===SubSECTION===
\subsection{Domain adaptation with diffusion-based fine-tuning}
% \textbf{Is optimizing data representations with generative loss enough to adapt the classifier?}
In the previous section, we evaluate whether the classifier can benefit from the generative part of our model when trained with limited access to labeled data. Now, we further extend those experiments and check if joint diffusion can adapt to the new data domain using only the generative part -- in a fully unsupervised way.
For this purpose, we run an experiment in which we first train the model on the source labeled data to retrain it on the target dataset without access to the labels. 
We compare our approach to a standalone deep neural network-based classifier, see Table \ref{tab:domain_adaptation}.

\begin{table}[htbp!]
  \vskip -4mm
  \centering
  \caption{The classification accuracy of the classifier trained in a domain adaption task. We first train the joint model on the source dataset, which we adapt to the target domain by retraining it using only the diffusion loss for the examples in the target one.}
  \medskip
  \resizebox{\linewidth}{!}{
  \def\arraystretch{1.25}
    \begin{tabular}{l|c|c|c}
        \hline
         \multicolumn{1}{c|}{} & SVHN $\rightarrow$ MNIST & USPS $\rightarrow$ MNIST & MNIST $\rightarrow$ USPS \\
        % \midrule
        \hline
        % Classifier & 95.14& 78.9 & 97 & 55 & 99 & 72\\
        Classifier & 78.8 & 54.7 & 72.2\\
        % \hdashline 
        % \makecell[l]{Joint diffusion} & 94.95 & 85.5 & 57 & 90 & 98.4 & 92.66 \\
        \textbf{Ours} & 85.5 & 90.5 & 92.7 \\
        % \midrule
        \midrule
        \makecell[l]{Classifier on target \\(upper bound)} & 96.1 & 96.8 & 99.4\\
        \hline
      \end{tabular}
  }
  \label{tab:domain_adaptation}
  \vskip -3.5mm
\end{table}

As expected, in all three scenarios, the classification accuracy of the stand-alone classifier degrades on a target domain.\footnote{The classification accuracy does not drop to a random level because all datasets share the same task, i.e., digits classification.} 
However, having access to unlabeled data from the target domain allows our joint diffusion model to adapt surprisingly well. Our approach outperforms the stand-alone classifier in all three cases by a significant margin. This result indicates that learning low-level features is essential for obtaining good predictive power while it is enough to transfer the classification head unchanged.

\section{Training details and hyperparameters}\label{app:training_details}

\subsection{Pooling of the UNet features}\label{app:pooling_details}
As discussed in Section~\ref{sec:diffusion_representations}, we pool the UNet features encoded to different UNet levels with the average pooling function. Precisely speaking, we take an average convolutional filter activation for a given filter across the whole image. This approach seems to result in the loss of information, such as the location of particular features extracted by the convolutional filter, but it allows us to create image representation with reasonable dimensionality. Depending on the dimensionality of input, with our method, we extract 1856 features for $28\times28$ Grayscale images (e.g. MNIST), 3712 features for $32 \times 32$ images with 3 color channels (e.g. CIFAR), and 5248 features for $64 \times 64$ images with 3 color channels (CelebA).

In all of our experiments, we use average pooling. Although other options such as max or min pooling might be used, our approach ensures that all of the features across the whole image are shared between the classifier and the generative models.

\subsection{Semi-supervised learning}\label{app:semi_supervised_training_details}
In our semi-supervised learning, we train our joint diffusion model on datasets with limited access to labeled samples. The simplest approach for this problem is to calculate the loss function on the diffusion using the whole batch of data while using only the labeled examples for the classifier loss. However, in some scenarios, we artificially omit up to 99\% of labeled data. In practice, this would lead to a situation where for batch size equal to 128 or 254 examples, the classifier loss would be practically calculated on 1 or 2 samples. Therefore, to stabilize the training we propose to create a buffer where we put labeled examples from each batch. When the buffer reaches its capacity equal to the batch size, we calculate the classifier loss using the examples from the buffer and add it to the generative loss according to Equation~\ref{eq:final_joint}.

\subsection{Domain adaptation}\label{app:domain_adaptation_training_details}
% \kamil{Describe fine-tuning}

In the experiments on the domain adaptation task, we propose the simplest setup, where we first train the joint model on the source task using the joint loss function (Eq.~\ref{eq:final_joint}), and then we retrain the model on the target domain using only the DDGM loss in Equation~\ref{eq:l_t_simple_ours}. We show that without any alteration to our basic setup, we can observe a significant performance boost compared to the baseline classifier. We believe that we can further improve those results if we focus directly on the domain adaptation task and take advantage of the recent advantages in this field. Further experiments in this direction should for example include simultaneous training on examples from two domains. To improve the alignment, we can also benefit from adversarial training as introduced by~\cite{ganin2016domain} in DANN.

%====SECTION====
\section{Additional results: Conditional generations with optimized representations}\label{app:conditional_generation}

\begin{figure*}[!htbp]
    \centering
    \begin{tabular}{cc}
    \includegraphics[width=.4\linewidth]{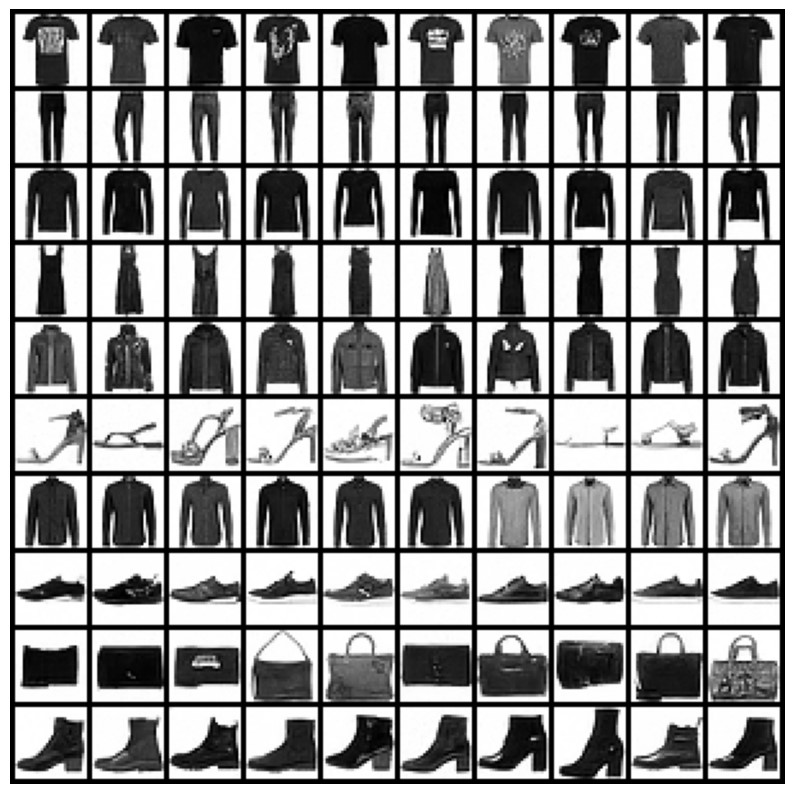} & \includegraphics[width=.4\linewidth]{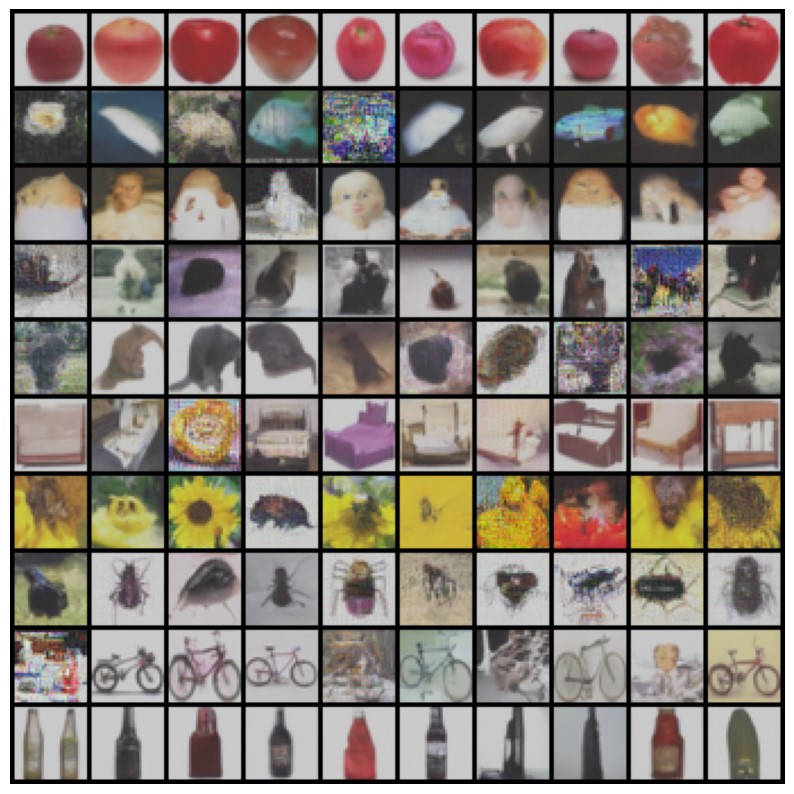}\\
    Fashion MNIST & CIFAR-100\\
    \end{tabular}
    \caption{Conditional samples from our joint diffusion model for Fashion MNIST dataset (\textit{left}) and first 10 classes of CIFAR100 dataset (\textit{right}). Each row represents samples from one class.}
    \label{fig:app_generations_additional}
\end{figure*}

\begin{figure*}[!htbp]
    \centering
    \begin{tabular}{ccc}
    \includegraphics[width=.3\linewidth]{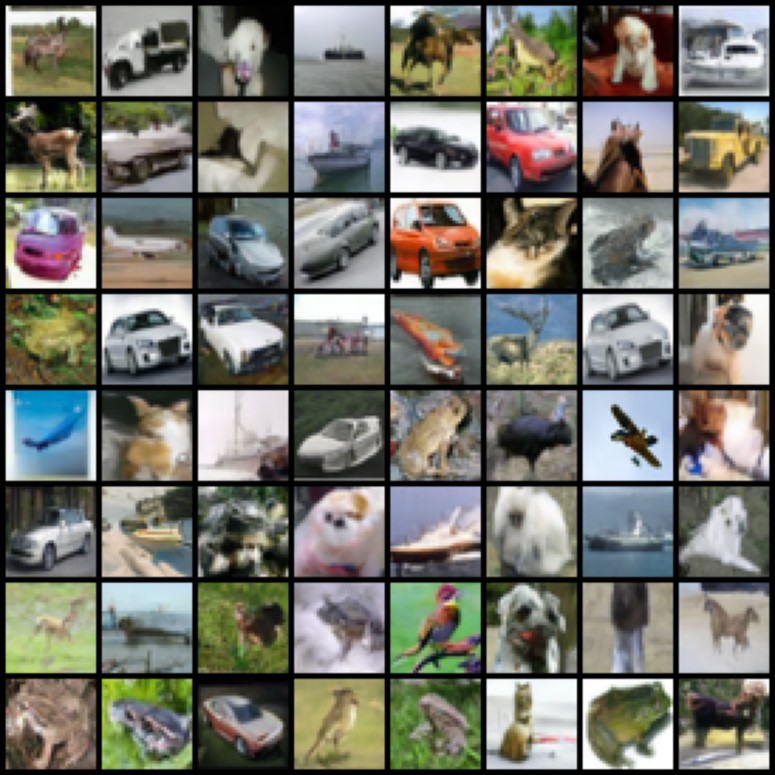} &
    \includegraphics[width=.3\linewidth]{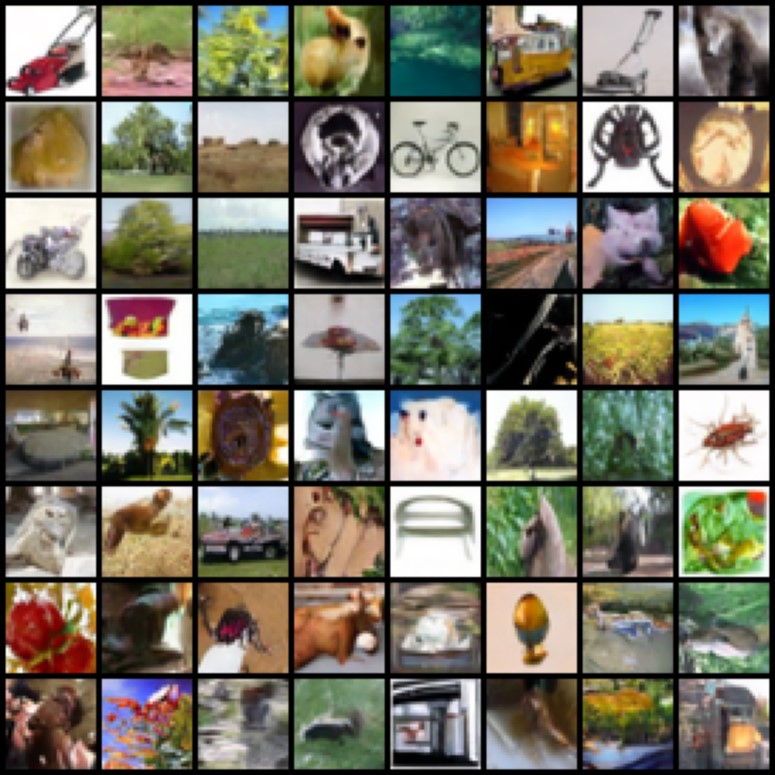} &\includegraphics[width=.3\linewidth]{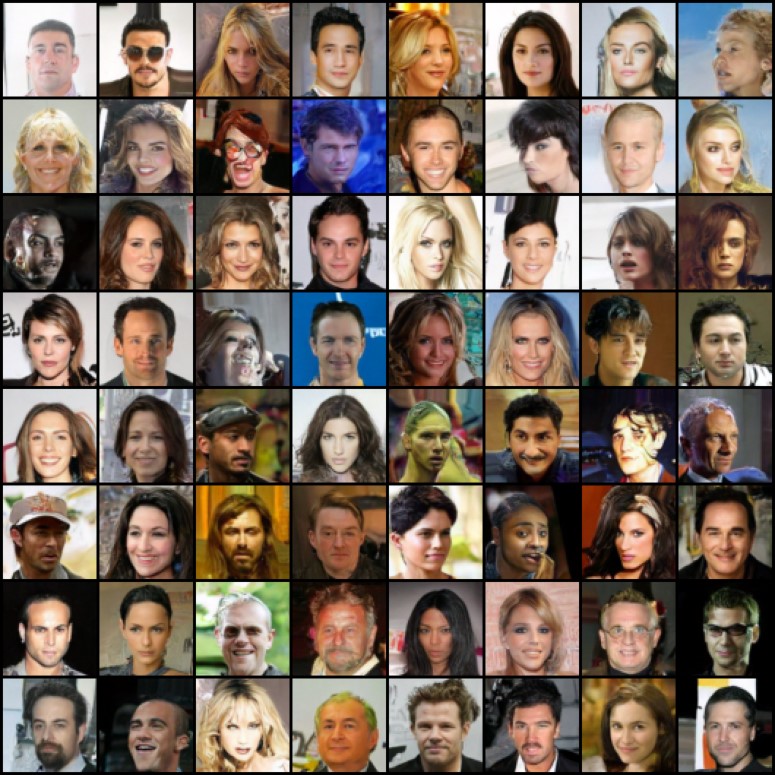}\\
    Fashion MNIST & CIFAR-100 & CelebA\\
    \end{tabular}
    \caption{Generated examples from our joint diffusion model without conditional sampling for CIFAR-10, CIFAR-100, and CelebA dataset.}
    \label{fig:app_generations_no_cond}
\end{figure*}

% \begin{figure}[h]
%     \centering
%     \begin{tabular}{c}
%     \includegraphics[width=.85\linewidth]{Figures/cifar_samples_200.png}\\
%     \makecell{$\alpha=200$}\\
%     % \includegraphics[width=.85\linewidth]{Figures/CIFAR10_classes_timesteps_500.pdf}\\
%     % $\alpha=500$\\
%     \includegraphics[width=.85\linewidth]{Figures/cifar_samples_1000.png}\\
%     $\alpha=1000$\\
%     \end{tabular}
%     \caption{CIFAR10: conditional samples with different values of step size $\alpha$}
%     \label{fig:app_cifar_samples}
% \end{figure}

\newpage
%====SECTION====
\section{Additional results: Counterfactual image generation}\label{app:counterfactual_generation}
\begin{figure*}[!htbp]
    \centering
    \begin{tabular}{c}
    \includegraphics[width=.6\linewidth]{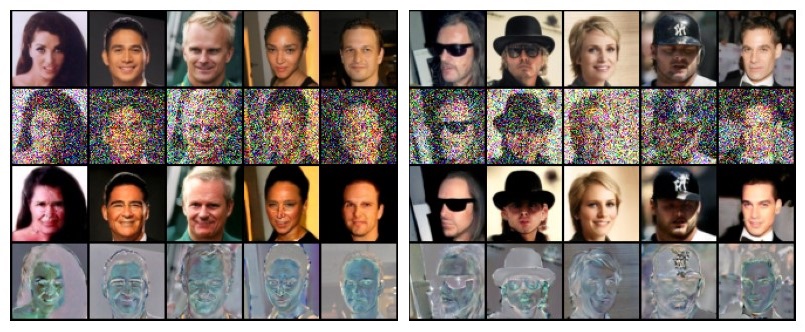}\\
    \makecell{\textit{(a)} Young to old (\textit{left}), old to young (\textit{right})}\\
     \\
    \includegraphics[width=.6\linewidth]{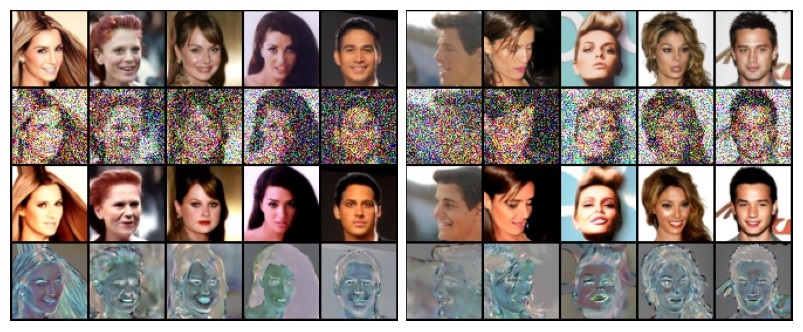}\\
    \textit{(b)} Smiling to no-smiling (\textit{left}), no-smiling to smiling (\textit{right})\\
     \\
    \includegraphics[width=.6\linewidth]{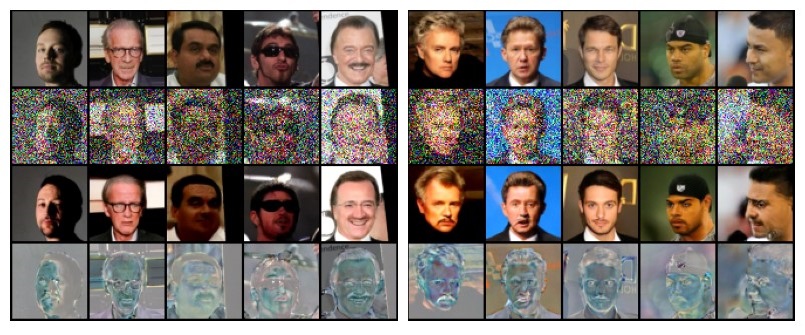}\\
    \textit{(c)} Mustache to no-mustache (\textit{left}), no-moustache to moustache (\textit{right})\\
    \end{tabular}
    \caption{Counterfactual image generation for the CelebA dataset using three different attributes on random original examples. For each attribute, we select 5 positive examples that we change to negative ones and 5 negative ones that we change to positive ones.}
    \label{fig:app_celeba_perturbations}
\end{figure*}

In the experiment described in Section~\ref{sec:medical}, we presented how we can use our joint diffusion model to generate the counterfactual explanations to the classifier using the medical dataset. In Figure~\ref{fig:app_celeba_perturbations}, we present more examples of this approach by perturbing original examples from the CelebA dataset. We select 3 attributes from the CelebA dataset namely: \emph{young}, \emph{smiling}, and \emph{mustache}. For each attribute, we select 5 positive examples and 5 negative examples which we alter using our conditional sampling procedure with the classifier-based optimization. We present original examples (first row) noised with 20\% of noise (second row) and generated towards counterfactual class (third row). In the last row, we show the differences between the original and modified examples.

\end{document}